\documentclass[10pt,twocolumn,letterpaper]{article}

\usepackage{cvpr}              

\usepackage{multirow, colortbl}

\definecolor{cvprblue}{rgb}{0.21,0.49,0.74}
\usepackage[pagebackref,breaklinks,colorlinks,allcolors=cvprblue]{hyperref}

\def\paperID{} 
\def\confName{CVPR}
\def\confYear{2026}

\title{View-Aware Semantic Alignment for Aerial-Ground Person Re-Identification}
    
\author{Quan Zhang$^{1*}$ {Zeqiang Cai$^{1*}$} Peiming Zhao$^1$ Jingze Wu$^1$ Cailun Wu$^1$
\\Hongbo Chen$^1$ Jianhuang Lai$^{1,2,3,4\dagger}$\\
$^1$ Sun Yat-sen University, China \\
$^2$ Pazhou Lab (HuangPu), Guangdong, China \\
$^3$ Guangdong Province Key Laboratory of Information Security Technology, Guangzhou, China \\
$^4$ Key Laboratory of Machine Intelligence and Advanced Computing, Ministry of Education, China \\
\tt\small\{zhangq689, wuclun, chenhongbo, stsljh\}@mail.sysu.edu.cn, \\
\tt\small\{caizq5, zhaopm, wujz3\}@mail2.sysu.edu.cn
}

\begin{document}
\maketitle
\let\thefootnote\relax
\footnotetext{$^*$ Equal contribution.}
\footnotetext{$^\dagger$ Corresponding author.}

\begin{abstract}
Aerial-Ground Person Re-Identification (AGPReID) remains highly challenging due to drastic viewpoint variations between drones and fixed cameras. Existing methods typically follow a view-invariant paradigm, aligning shared features across views to achieve robustness. However, view-invariant inherently enforces part-level alignment, which ignores view-specific cues and discriminative identity information. To this end, this work proposes ViSA (View-aware Semantic Alignment), a view-aware framework that achieves cross-view semantic consistency containing an Expert-driven Token Generation Module (ETGM) and a Dual-branch Local Fusion Module (DLFM). Technically, the former constructs a set of view-aware experts to generate adaptive semantic queries that perceive viewpoint-specific patterns, while the latter leverages graph reasoning to extract and align local regions responsive to different experts. Extensive experiments on three AGPReID benchmarks including AG-ReID.v2, CARGO and LAGPeR demonstrate that ViSA consistently achieves superior performance, with a notable 10.06\% mAP improvement on the challenging CARGO cross-view protocol. The code is available at \href{https://github.com/Cat-Zero/ViSA}{https://github.com/Cat-Zero/ViSA}.
\end{abstract}
    
\vspace{-3mm}
\section{Introduction}
\label{sec:intro}

Person Re-Identification (ReID) aims to match pedestrian images of the same individual across non-overlapping cameras. While conventional ReID models have achieved remarkable progress under single view settings~\cite{zhang2024separable, wang2025idea, zhang2024uncertainty, liu2023video, zhang2021seeing, lu2023learning, zhang2022modeling, shi2024multi, gao2024part, peng2023feature, zhang2021learning, 10318082, zhang2021hat, NEURIPS2023_7df69dbf}, their assumptions break down when facing drastic viewpoint disparities introduced by aerial and ground sensors. The integration of drones into surveillance networks gives rise to the Aerial-Ground Person Re-Identification (AGPReID) problem~\cite{nguyen2023agreid, zhang2020person, nguyen2024agreidv2, zhang2023ground}, which involves extreme viewpoint variations.

\begin{figure}[t]
    \centering
    \includegraphics[width=\linewidth]{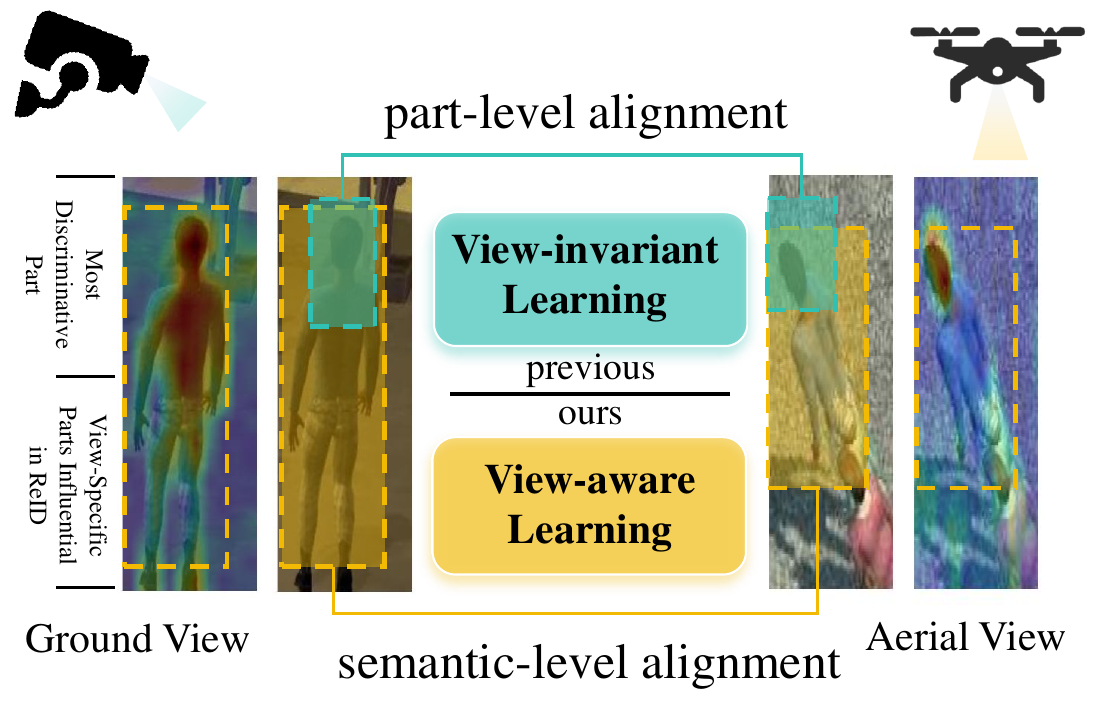}
    \caption{Comparison of view-invariant and view-aware learning. View-invariant learning typically converges to part-level alignment, whereas view-aware learning achieves semantic-level perception for more robust cross-view representations.}
    \label{fig:intro}
\end{figure}

Most existing AGPReID methods follow a view-invariant paradigm that aligns shared representations across heterogeneous viewpoints~\cite{zhang2024view, sun2019dissecting, wang2025secap, khalid2025bridging}. As illustrated in Fig.\ref{fig:intro}, although effective to some extent, this paradigm enforces excessive cross-view alignment, resulting in shared-part matching and inadvertently suppressing discriminative, view-specific cues that are closely correlated with identity such as leg under ground view and shoulder under aerial view. Moreover, existing methods predominantly rely on global representations, overlooking that fine-grained identity evidence often resides in local patch regions. In aerial imagery, steep downward angles and occlusions often cause the loss of partial body parts, which are highly discriminative in ground-view images, further challenging the reliability of global descriptors. These limitations motivate a paradigm shift. Instead of suppressing viewpoint factors, an effective AGPReID framework should achieve cross-view semantic consistency by disentangling viewpoint variations while retaining and leveraging view-specific identity cues. 

To this end, we propose \textbf{ViSA} (\textbf{Vi}ew-aware \textbf{S}emantic \textbf{A}lignment), a view-aware framework designed to jointly capture view-invariant identity cues and view-specific discriminative features. At the core of ViSA are two complementary modules that operate on top of a view-aware backbone. First, the {Expert-driven Token Generation Module (ETGM)} constructs a set of learnable, view-aware expert tokens from both the view-invariant and view-specific representations produced by the encoder. Each expert specializes in capturing distinct identity patterns that may manifest differently under aerial or ground perspectives. By employing a mixture-of-experts mechanism, ETGM adaptively routes information from local patch tokens into semantic queries that preserve complementary identity cues while maintaining separation between invariant and viewpoint-dependent components. The {Dual-branch Local Fusion Module (DLFM)} extracts and aligns discriminative local features in a structured manner. Specifically, for each expert query, DLFM selects a sparse set of the most semantically relevant local patches using cosine similarity, forming a query-guided neighborhood graph. A graph convolutional network then operates on this query-augmented local graph to capture structural relationships among the selected patches, effectively embedding each expert into its responsive local regions. By maintaining separate branches for view-invariant and view-specific queries, DLFM ensures that both robust identity information and systematic viewpoint variations are preserved. Finally, the refined local features from both branches are concatenated with the global [CLS] token, producing a comprehensive and discriminative representation suitable for cross-view retrieval. 

In summary, our main contributions are as follows:
\begin{itemize}
\item We propose a view-aware semantic alignment paradigm for AGPReID, which learns to disentangle and jointly exploit both view-invariant and view-specific cues. This design enables more comprehensive and robust identity modeling across highly heterogeneous viewpoints.
\item We design ViSA (\textbf{Vi}ew-aware \textbf{S}emantic \textbf{A}lignment), a novel framework that integrates an Expert-driven Token Generation Module (ETGM) for adaptive view-aware query construction and a Dual-branch Local Fusion Module (DLFM) for expert-guided local feature alignment across aerial and ground views.
\item We conduct extensive experiments on three large-scale benchmarks including AG-ReID.v2, CARGO, and LAGPeR, demonstrating that ViSA consistently surpasses state-of-the-art methods and achieves up to a 10.06\% mAP improvement under cross-view protocol.
\end{itemize}

\section{Related Work}
\label{sec:formatting}

\subsection{Aerial-Ground Person ReID}

With the rapid development of UAV-based imaging technology, AGPReID has attracted increasing attention. Nguyen et al. \cite{nguyen2023agreid} first introduced the AG-ReID.v1 dataset and proposed a dual-stream interpretable network that leverages attribute information to mitigate the challenges of cross-view retrieval. They later extended this dataset to AG-ReID.v2 \cite{nguyen2024agreidv2}, emphasizing head-region features to further improve cross-view matching performance. Hu et al. \cite{zhang2025latex} adopted prompt-tuning strategies to incorporate attribute-based textual knowledge into model training, enabling more effective utilization of semantic information.

Unlike attribute-based approaches, which require expensive manual annotations, VDT \cite{zhang2024view} explored a view-invariant paradigm from the perspective of view decoupling, aiming to capture shared representations across views. Building upon this paradigm, several subsequent works further advanced its performance: DTST \cite{awang2024dynamictokenselectionaerialground} proposed dynamic token selection to enhance feature representation, and SeCap \cite{wang2025secap} leveraged prompt learning to capture local features. Recently, VIF-AGReID \cite{khalid2025bridging} proposed patch-level rotational augmentation and an angular-penalty loss to strengthen view-invariant feature learning in aerial-ground scenarios. Despite their effectiveness, these methods fundamentally rely on shared-part alignment, which enforces the alignment of view-specific parts and consequently limits the model’s ability to capture full-body pedestrian information. This limitation motivates our proposed view-aware semantic alignment, which aims to preserve view-invariance while fully exploiting global information.

\subsection{Mixture of Experts in ReID}

Mixture-of-Experts (MoE) models \cite{jacobs1991adaptive, jordan1994hierarchical} consist of multiple expert networks coordinated by a gating network, which dynamically selects the most suitable expert to process a given input. MoE has demonstrated remarkable effectiveness across various fields, including computer vision \cite{hwang2023tutel, riquelme2021scaling, chowdhury2023patch} and multimodal learning \cite{chen2024llava, chen2023octavius, mustafa2022multimodal}. In ReID, MoE architectures have also shown promising potential. MoSCE-ReID \cite{REN2025129587} proposed a MoE architecture specifically designed for distinct attribute groups, effectively addressing inherent attribute conflicts and generalization challenges in ReID. HAMoBE \cite{Su_2025_ICCV} mimics human perceptual mechanisms by independently modeling key biometric features and adaptively integrating them. Additionally, numerous other works \cite{Wang_Liu_Zheng_Zhang_2025,wang2025moda} have explored various MoE variants in ReID to enhance feature expressiveness and generalization.

\begin{figure*}[t]
    \centering
    \includegraphics[width=\linewidth]{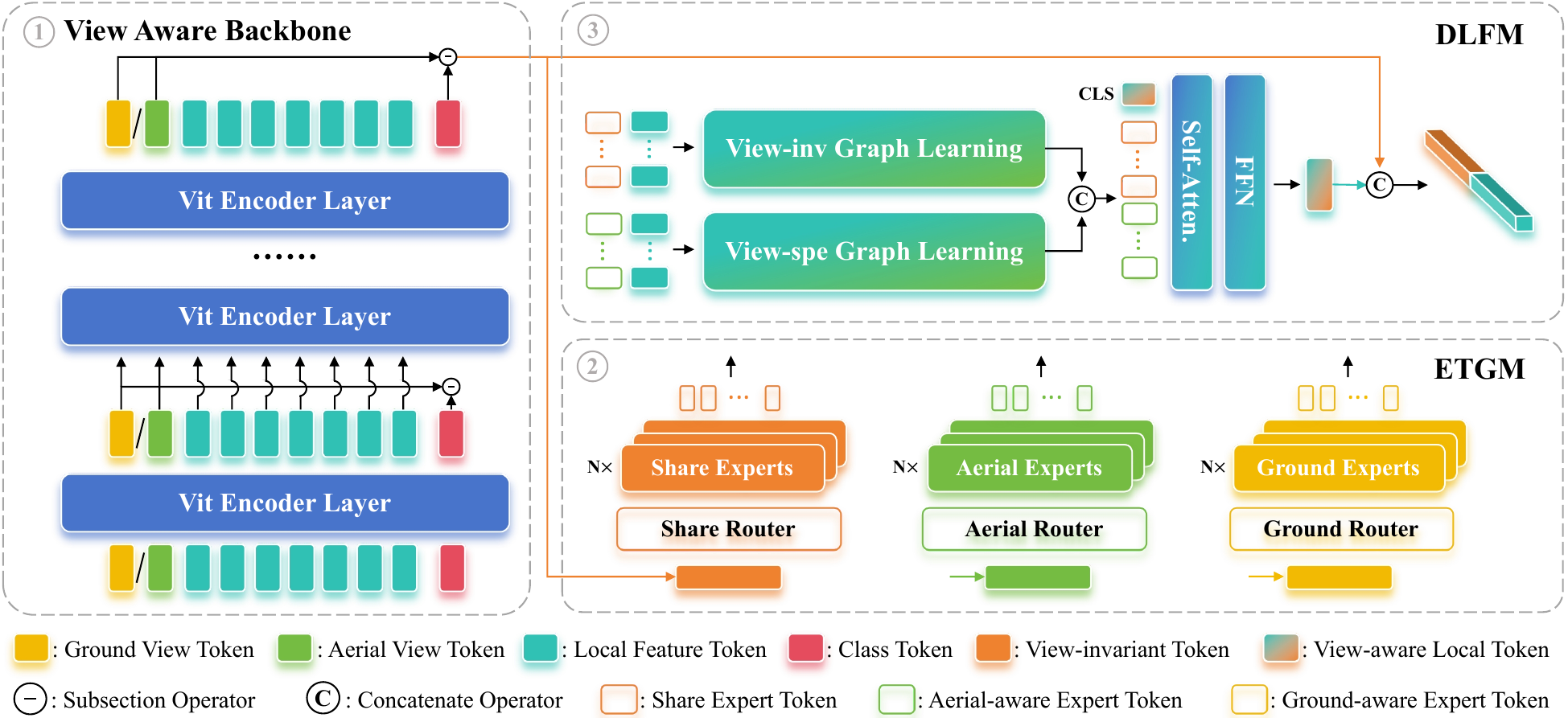}
    \caption{Overview of the proposed ViSA. The transformer backbone first separates intrinsic identity semantics from extrinsic viewpoint factors using dual aerial/ground tokens. Then the Expert-driven Token Generation Module (ETGM) employs a mixture-of-experts mechanism including shared, aerial, and ground experts to generate adaptive semantic tokens guided by corresponding routers. Finally the Dual-branch Local Fusion Module (DLFM) performs graph-based reasoning to decode view-invariant and view-specific local features. View-inv Graph Learning and View-spe Graph Learning represents the view-invariant branch and view-specific branch respectively.}
    \label{fig:model}
\end{figure*}
Despite the significant progress of existing MoE methods, the design of experts specifically targeting view variations remains largely unexplored. This observation motivates our proposed view-specific experts, which allocate dedicated experts for different viewpoints, enabling more effective modeling of cross-view feature discrepancies while preserving global pedestrian information.

\subsection{Graph Convolutional Networks in ReID}

Graph Convolutional Networks (GCNs) are capable of propagating and aggregating features over graph-structured data and have received increasing attention in ReID tasks in recent years. GPS \cite{nguyen2021graph} constructed graphs from pedestrian attribute labels and visual features, leveraging GCNs to learn the topological structure of the human body and integrating it into the ReID framework to enhance feature representations. To address occluded pedestrian scenarios, ADGC \cite{wang2020high} proposed utilizing topological information and higher-order relationships to achieve robust alignment, while RTGAT \cite{huang2023reasoning} designed an inference-and-refinement graph attention network to learn complete representations from occluded images. Inspired by these works, our method leverages GCNs to capture view-aware topological information from local features, enabling cross-view semantic alignment and more robust pedestrian representations.
\section{Method}

\begin{figure*}[ht!]
    \centering
    \includegraphics[width=\linewidth]{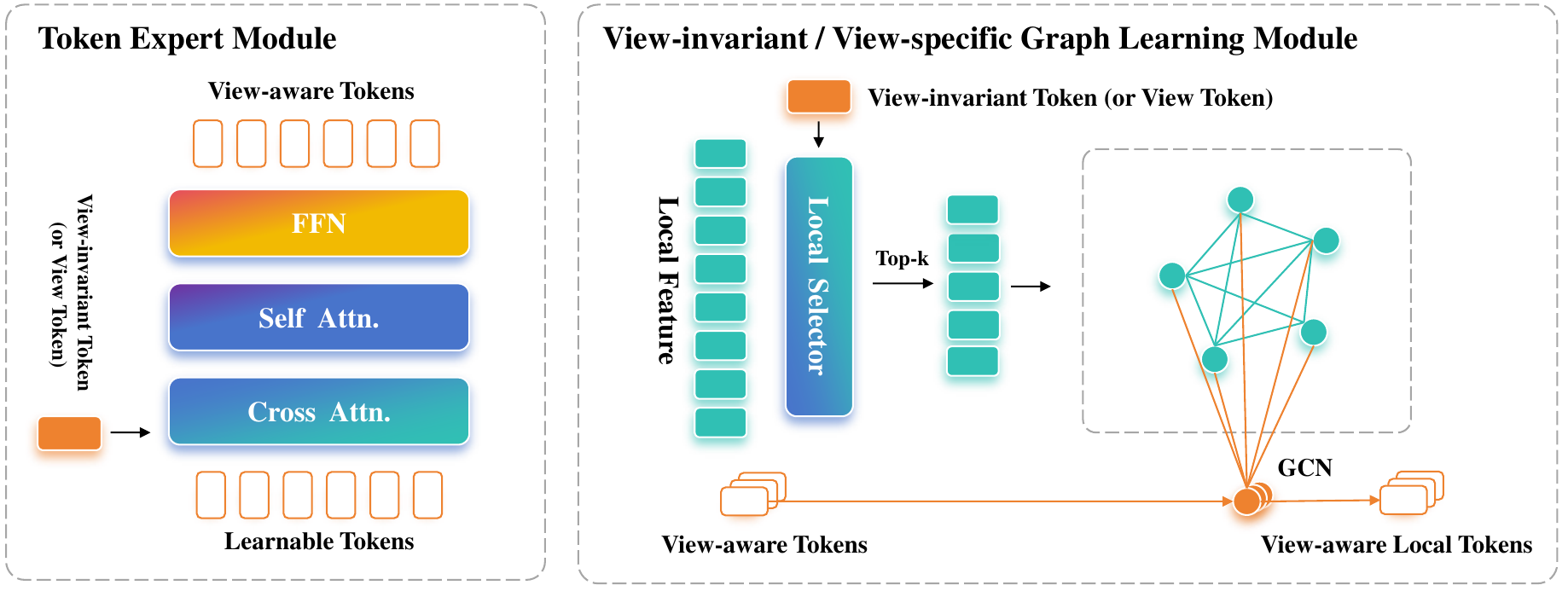}
    \caption{Modules detail of the Token Expert Module and Graph Learning Module. The Token Expert Module generates view-aware expert tokens from view tokens and view-invariant tokens. The Graph Learning Module fuses view-aware tokens with a patch-level graph via graph convolution to refine local cues and enhance cross-view ReID representations.}
    \label{fig:model_detail}
\end{figure*}

\subsection{Problem Statement}
In cross-view person re-identification, the visual observation $I$ of a person is jointly determined by intrinsic identity factors $X$ (e.g., body shape, appearance) and extrinsic viewpoint factors $V$ (e.g., camera angle, height, and perspective distortion). The feature representation $Z$ extracted by a neural encoder therefore inevitably entangles these two underlying sources of variation:
\begin{align}
    Z = f(I) = f(X, V).
\end{align}
Such entanglement makes it challenging to separate identity cues from viewpoint variations, especially under drastic aerial-ground changes. When the viewpoint $V$ changes significantly, the same identity produces very different observations, making $Z$ highly view-biased. Traditional methods suppress the influence of $V$ to learn view-invariant features, but such suppression often removes useful discriminative cues correlated with identity, leading to information loss.

To formally describe this trade-off between invariance and discriminability, we adopt an information-theoretic perspective, where the representation should contain sufficient information for identity recognition while minimizing irrelevant dependencies on viewpoint:
\begin{align}
    \max I(Z;Y) \ \  \text{s.t.} \ \ I(Z;V)\ \text{is minimized,}
\end{align}
where $Y$ denotes the person identity label. However, strictly minimizing $I(Z;V)$ may also reduce $I(Z;Y)$, because some view-related cues (e.g., shape consistency, clothing folds) are partially correlated with identity. To balance this trade-off, we adopt a disentangled representation:
\begin{align}
    Z = [Z_{inv}, Z_{spe}], I(Z_{inv};V)\approx0, I(Z_{spe};V)>0,
\end{align}
where $Z_{inv}$ preserves identity-relevant, view-invariant semantics, while $Z_{spe}$ explicitly encodes the systematic appearance variations induced by viewpoint. Such decomposition avoids the information bottleneck caused by adversarial suppression and allows the model to exploit complementary cues between $Z_{inv}$ and $Z_{spe}$. This theoretical insight directly inspires our framework design, where the encoder explicitly decouples $Z_{inv}$ and $Z_{spe}$, and the decoder hierarchically re-fuses them to robustly preserve both fine-grained identity and viewpoint information.

\subsection{Framework}

The overall architecture of ViSA is illustrated in Fig. \ref{fig:model}, following an encoder-decoder design. The encoder extracts multi-view semantic representations, while the decoder explicitly disentangles and aggregates fine-grained identity cues to produce discriminative cross-view features.

The proposed ViSA is built upon the View-Decoupled Transformer (VDT)~\cite{zhang2024view} and extend its design to a dual-stream encoder that preserves both invariant and viewpoint-sensitive semantics. Each layer introduces independent learnable tokens for aerial and ground views, allowing the model to capture viewpoint characteristics. Following VDT, the [CLS] token progressively subtracts its corresponding view token, removing spurious view bias and forming a more view-invariant and holistic global representation.

However, the [CLS] token alone cannot capture distributed identity cues such as clothing textures, body poses, and local semantics, which are intertwined with viewpoint variations. To address this limitation, we introduce two complementary modules in the decoder. The Expert-driven Token Generation Module (ETGM) performs viewpoint-aware query routing through a mixture-of-experts mechanism, while the Dual-branch Local Fusion Module (DLFM) employs graph reasoning to refine patch-level representations. Together, they recover fine-grained identity cues while maintaining explicit modeling of viewpoint-dependent variations. The final feature representation concatenates refined local features with the [CLS] token, yielding a robust and discriminative embedding for aerial-ground re-identification.



\subsection{Expert-driven Token Generation Module}

To exploit the distributed identity information in local patch tokens while handling viewpoint variations, we propose the Expert-driven Token Generation Module (ETGM), which produces a set of semantic-guided query tokens to disentangle view-invariant and view-related features. Formally, given the encoder outputs $Z_{inv}$ and $Z_{spe}$, ETGM generates queries $Q_{inv}$ and $Q_{spe}$ satisfying:
\begin{align}
    &Q_{inv} = \text{ETGM}(Z_{inv}), I(Q_{inv};Y)\ \text{is minimized,}\\
    &Q_{spe} = \text{ETGM}(Z_{spe}), \\
    &I(Q_{inv}; V)\approx 0, I(Q_{spe};Y|V)>0,
\end{align}
where $Y$ denotes identity labels and $V$ denotes viewpoints. As shown in Fig.\ref{fig:model_detail}, each expert in the MoE consists of a learnable token set $\{t_1,\dots,t_M\}$, which interacts with the input features through a transformer block comprising cross-attention, self-attention, and feed-forward layers:
\begin{align}
    T'=\text{FFN}(\text{SelfAttn}(\text{CrossAttn}(T,Z))),
\end{align}
where $T$ are the expert tokens and $Z$ is the input feature ($Z_{inv}$ or $Z_{spe}$). Cross-attention allows the tokens to integrate information from the input feature, self-attention enables interaction between tokens, and the FFN provides a non-linear transformation. The outputs of multiple selected experts are aggregated via a weighted sum to form the final query tokens $Q_{inv}$ and $Q_{spe}$.

From an information-theoretic perspective, this design maximizes the mutual information between the query and the identity label while explicitly preserving conditional view-dependent information. By separating view-invariant and view-related queries, ETGM provides flexible, distributed guidance for downstream local feature aggregation in the DLFM module, ensuring that both core identity cues and auxiliary view-dependent identity signals are retained.

\subsection{Dual-branch Local Fusion Module}

The Dual-branch Local Fusion Module (DLFM) aims to exploit the distributed identity information embedded in local patch tokens while explicitly modeling viewpoint dependencies. 
Guided by the expert queries generated by ETGM, DLFM contains a view-invariant branch and a view-related branch, each equipped with a Graph Convolutional Network to capture structural relations among local tokens.

The detail of graph learning module used in DLFM is shown in Fig.\ref{fig:model_detail}.  Formally, given the local patch feature set $F=\{f_i\}_{i=1}^N$ from the encoder and the expert queries $Q_{inv}$ and $Q_{spe}$, a naive approach would directly apply cross-attention between $Q$ and $F$. 
However, such attention neglects the distributed structural information inherently encoded within local patches. 
To overcome this, each query first selects the top-$K$ most semantically relevant local tokens based on cosine similarity:
\begin{align}
\mathcal{N}_k(Z) = \text{TopK}(\cos(Z,F)),
\end{align}
where
\begin{align}
\cos(i, j) = \frac{\boldsymbol{x}_i^\top \boldsymbol{x}_j}{\|\boldsymbol{x}_i\|_2 \|\boldsymbol{x}_j\|_2}, 
\quad i,j \in \{1, \dots, N\}.
\end{align}

This sparse neighborhood selection enforces semantic locality, allowing the model to focus on regions that share consistent semantics with the query while suppressing irrelevant patches. 
Based on the selected neighbors, a fully connected local graph $\mathcal{G}=\{\mathcal{N}_k(Z), A(Z)\}$ is constructed, where each edge weight encodes pairwise cosine similarity:
\begin{align}
A_{ij}(Z) = \cos(f_i,f_j),\quad f_i, f_j\in \mathcal{N}_k(Z).
\end{align}

To further inject query semantics into local reasoning, the graph is augmented by inserting the query token $Q$ as an additional node, forming an expanded node set $\mathcal{N}_{qv}(Q) = [Q, \mathcal{N}_k(Z)]$. 
The corresponding edge matrices for the invariant and view-related branches are defined as:
\begin{align}
\mathbf{A}_{qv}(Q) =
\begin{bmatrix}
1 & \text{sim}(Q, \mathcal{N}_k(Z)) \\
\text{sim}(\mathcal{N}_k(Z), Q) & \mathcal{N}_k(Z)
\end{bmatrix}.
\end{align}

This query-augmented graph enables the model to embed each query into the local feature manifold, integrating richer structural and semantic relations from the most relevant local regions.
We then apply a dedicated GCN $g_{qv}$ to adapt $Q$ within each subgraph:
\begin{align}
\mathcal{N}^*_{qv}(Q) = g_{qv}(\mathcal{N}_{qv}(Q), \hat{A}_{qv}(Q)),
\end{align}
where $\hat{A}$ denotes the Laplacian-normalized adjacency matrix. 
The refined queries $\hat{Q} = \mathcal{N}^*_{qv}(Q)[0, :]$ from both branches are concatenated and fused with the global [CLS] token via a self-attention block:
\begin{align}
F_{local} = \text{Attn}([\hat{Q}_{inv}, \hat{Q}_{spe}, \text{CLS}]),
\end{align}
yielding the final local representation.

By combining semantic sparsification (Top-$k$ selection) with dual-branch disentanglement, DLFM jointly captures identity-consistent and view-dependent cues in a unified representation. 
The invariant branch provides robust identity-discriminative features, while the view-related branch models systematic viewpoint-dependent variations, thereby enhancing overall cross-view discriminability.

\subsection{Optimization}

Our model is jointly optimized with multiple objectives that supervise identity learning, viewpoint disentanglement, and expert utilization. These joint objectives together ensure that ViSA learns highly identity-discriminative yet viewpoint-adaptive feature representations.

\noindent\textbf{Identity Supervision.}  
To learn discriminative and compact identity embeddings, we employ the standard cross-entropy loss and triplet loss for each $x_i \in B$:
\begin{gather}
\mathcal{L}_{id} = -\frac{1}{|B|}\sum_{i=1}^{|B|}\log p(y_i|f_i), \tag{14}\\
\mathcal{L}_{tri} = \frac{1}{|B|}\sum_{i=1}^{|B|}\left[||f_i, f_i^+||_2^2 - ||f_i, f_i^-||_2^2 + m\right]_+. \tag{15}
\end{gather}
where $|B|$ is the batch size, $f_i$ denotes the global or refined local feature of the $i$-th sample, and $m$ is the margin.

\begin{table*}[!htb]
    \centering
    \caption{Performance comparison on CARGO dataset. `ALL' denotes the overall retrieval performance. `G$\leftrightarrow$G', `A$\leftrightarrow$A', `A$\leftrightarrow$G' represents the performance of each model in several specific retrieval patterns. The best and second-best results are highlighted in \textbf{bold} and \underline{underline}, respectively for clear comparative emphasis.}
        \begin{tabular}{lccccccccc}
        \toprule
        \multirow{2}{*}{Method} & \multirow{2}{*}{Venue} & \multicolumn{2}{c}{ALL} & \multicolumn{2}{c}{A$\leftrightarrow$G} & \multicolumn{2}{c}{G$\leftrightarrow$G} & \multicolumn{2}{c}{A$\leftrightarrow$A}  \\
        \cmidrule(lr){3-4}  \cmidrule(lr){5-6} \cmidrule(lr){7-8} \cmidrule(lr){9-10}
        & & Rank-1 & mAP & Rank-1 & mAP & Rank-1 & mAP & Rank-1 & mAP \\
        \midrule
        MGN\cite{2018Learning} & ACM MM'18 & 54.81 & 49.08 & 31.87 & 33.47 & 83.93 & 71.05 & 65.00 & 52.96 \\
        BoT\cite{luo2019bag} & CVPR'19 & 54.81 & 46.49 & 36.25 & 32.56 & 77.68 & 66.47 & 65.00 & 49.79 \\
        VV\cite{kumar2020strong} & JAISCR'20 & 45.83 & 38.84 & 31.25 & 29.00 & 72.31 & 62.99 & 67.50 & 49.73 \\
        PCB\cite{2021Learning} & TPAMI'21 & 51.00 & 44.50 & 34.40 & 30.40 & 74.10 & 67.60 & 55.00 & 44.60 \\
        AGW\cite{ye2021Deep} & TPAMI'21 & 60.26 & 53.44 & 43.57 & 40.90 & 81.25 & 71.66 & 67.50 & 56.48 \\
        SBS\cite{he2023fastreid} & ACM MM'23 & 50.32 & 43.09 & 31.25 & 29.00 &72.31 & 62.99 & 67.50 & 49.73 \\
        \midrule
        ViT\cite{dosovitskiy2020image} & ICLR'21 & 61.54 & 53.54 & 43.13 & 40.11 & 82.14 & 71.34 & \underline{80.00} & 64.47 \\
        PCL-CLIP\cite{li2023prototypical} & CoRR'23 & 67.31 & 60.93 & 54.37 & 51.43 & 84.82 & 76.00 & 70.00 & 60.75 \\
        CLIP-ReID\cite{li2023clip} & AAAI'23 & 68.27 & \underline{64.25} & 55.62 & 53.83 & 84.82 & \underline{80.80} & 75.00 & 65.42 \\
        \midrule
        VDT\cite{zhang2024view} & CVPR'24 & 64.10 & 55.20 & 48.12 & 42.76 & 82.14 & 71.59 & \textbf{82.50} & 66.83 \\
        DTST\cite{awang2024dynamictokenselectionaerialground} & ICME'25 & 64.42 & 55.73 & 50.63 & 43.39 & 78.57 & 72.40 & \underline{80.00} & 63.31 \\
        SeCap\cite{wang2025secap} & CVPR'25 & \underline{68.59} & 60.19 & \underline{69.43} & \underline{58.94} & \underline{86.61} & 75.42 & \underline{80.00} & \textbf{68.08} \\
        \midrule
        \rowcolor{gray!15}
       ViSA & - & \textbf{70.51} & \textbf{65.46} & \textbf{71.28} & \textbf{69.00} & \textbf{88.39} & \textbf{83.90} & \textbf{82.50} &  \underline{67.78} \\
        \bottomrule    
        \end{tabular}   
    \label{tab:cargo_full}
\end{table*}

\noindent\textbf{Viewpoint Classification and Disentanglement.}  
To explicitly capture viewpoint information, a lightweight binary classifier predicts the camera domain (ground vs. aerial) from the view token.  
The corresponding classification loss is then formulated as:
\begin{align}
\mathcal{L}_{view} = -\frac{1}{|B|}\sum_{i=1}^{|B|} v_i \log \hat{v}_i, \tag{16}
\end{align}
where $v_i$ and $\hat{v}_i$ represent the ground-truth and predicted view labels, respectively.  
To disentangle identity and viewpoint representations more effectively, we impose an explicit orthogonality constraint between the invariant feature $f_i^{inv}$ and the view feature $f_i^{spe}$:
\begin{align}
\mathcal{L}_o = \frac{1}{|B|}\sum_{i=1}^{|B|}\left|\cos(f_i^{inv}, f_i^{spe})\right|. \tag{17}
\end{align}

\noindent\textbf{MoE Load-Balancing Loss.}  
For the ETGM, we adopt a load-balancing regularization to prevent expert collapse and encourage diverse expert participation:
\begin{gather}
\mathcal{L}_{balance} = E\sum_{j=1}^{E}\bar{p}_j^2, \tag{18}\\
\bar{p}_j = \frac{1}{|B|}\sum_{i=1}^{|B|}p_{f_i,j}, \tag{19}
\end{gather}
where E represents the number of experts and $p_{f_i,j}$ denotes the routing probability of the $j$-th expert for instance $i$.  
Minimizing this term promotes uniform expert utilization and stabilizes MoE training.

\noindent\textbf{Overall Objective.} 
The total optimization objective of ViSA can be formally written as:
\begin{align}
\mathcal{L} &= (\mathcal{L}_{id}^{global} + \mathcal{L}_{tri}^{global}) \notag\\
&+ (\mathcal{L}_{id}^{local} + \mathcal{L}_{tri}^{local}) \tag{20}\\
&+ \mathcal{L}_o + \mathcal{L}_{view} + \lambda\mathcal{L}_{balance}, \notag
\end{align}
where $\lambda$ is a weighting hyperparameter to balance the MoE load balancing and ReID objectives.  

\section{Experiment}

\subsection{Experiment Settings}

\begin{table*}[!htb]
    \centering
    \caption{The efficacy of components in ViSA is evaluated on the CARGO dataset. VAB represents using the view aware backbone. The best results are highlighted in \textbf{bold}.}
        \begin{tabular}{ccccccccccccc}
        \toprule
        \multirow{2}{*}{VAB} & \multirow{2}{*}{ETGM} & \multirow{2}{*}{DLFM} 
        & \multicolumn{2}{c}{ALL} 
        & \multicolumn{2}{c}{A$\leftrightarrow$G} 
        & \multicolumn{2}{c}{G$\leftrightarrow$G} 
        & \multicolumn{2}{c}{A$\leftrightarrow$A} 
        & \multicolumn{2}{c}{Average}  \\
        \cmidrule(lr){4-5}  \cmidrule(lr){6-7} \cmidrule(lr){8-9} \cmidrule(lr){10-11} \cmidrule(lr){12-13}
        & & & Rank-1 & mAP & Rank-1 & mAP & Rank-1 & mAP & Rank-1 & mAP & Rank-1 & mAP \\
        \midrule
        \midrule
        \multicolumn{3}{c}{Baseline (ViT)} 
        & 61.54 & 53.54 & 43.13 & 40.11 & 82.14 & 71.34 & {80.00} & 64.47 & 66.70 & 57.37 \\
        \midrule
        \checkmark & & & 64.10 & 55.20 & 48.12 & 42.76 & 82.14 & 71.59 & \textbf{82.50} & {66.83} & 69.22 & 59.10 \\
        & & \checkmark 
        & 67.31 & 62.86 & 65.96 & 66.72 & 86.61 & 81.85 & {80.00} & 66.44 & 74.97 & 69.47 \\
        \midrule
        \checkmark & & \checkmark 
        & 68.59 & 62.40 & 68.09 & 65.53 & 84.82 & 80.80 & {80.00} & 64.87 & 75.38 & 68.43 \\
        & \checkmark & \checkmark 
        & 69.55 & 64.06 & 69.15 & 66.59 & 83.04 & 80.85 & 75.00 & 65.18 & 74.19 & 69.17 \\
        \midrule
        \rowcolor{gray!15}
        \checkmark & \checkmark & \checkmark 
        & \textbf{70.51} & \textbf{65.46} & \textbf{71.28} & \textbf{69.00} & \textbf{88.39} & \textbf{83.90} & \textbf{82.50} &  \textbf{67.78} & \textbf{78.17} & \textbf{71.54} \\
        \bottomrule    
    \end{tabular}  
    \label{tab:cargo_ab}
\end{table*}

\noindent\textbf{Datasets.} We conduct extensive evaluations on three representative AGPReID benchmarks, including two real-world datasets and one synthetic dataset: AG-ReID.v2 \cite{nguyen2024agreidv2}, LAGPeR \cite{wang2025secap} and CARGO \cite{zhang2024view}. AG-ReID.v2 is a large-scale real-world dataset specifically designed for person Re-ID in mixed aerial–ground scenarios. It comprises 100,502 images of 1,615 unique identities, each annotated with 15 soft attribute labels. The data were collected from multiple sensing platforms, including UAVs, stationary CCTVs, and smart glasses–integrated cameras, covering diverse viewing angles and motion patterns. CARGO is a large-scale synthetic dataset comprising 108,563 person images from 5,000 identities, captured by eight ground and five aerial cameras, providing a controlled setting with diverse cross-view variations. LAGPeR contains 63,841 images covering 4,231 identities acquired by fourteen ground and seven aerial cameras, with drone altitudes between 20 m and 60 m, providing richer spatial diversity.

\noindent\textbf{Evaluation Metrics.} Following previous works, we adopt the cumulative matching characteristic (CMC) at Rank-1 (R1) and the mean Average Precision (mAP) as evaluation metrics. These complementary metrics jointly measure retrieval precision and ranking accuracy, providing a holistic evaluation of cross-view person retrieval performance.

\noindent\textbf{Implementation Details.} All experiments are conducted on a single NVIDIA RTX 4090 GPU. We adopt the Vision Transformer (ViT) backbone \cite{liu2021swin}, pre-trained on ImageNet \cite{deng2009imagenet}, as the feature encoder. Input images are uniformly resized to $256 \times 128$. During training, each mini-batch contains 256 samples, composed of 64 identities with 4 instances per identity. The model is optimized for 120 epochs using Stochastic Gradient Descent (SGD) \cite{bottou2010large} with momentum, starting from an initial learning rate of $8 \times 10^{-3}$ and decayed to $1.6 \times 10^{-6}$ following a cosine annealing schedule. For the ETGM, we adopt a mixture-of-experts architecture containing 8 experts for each type (view-invariant and view-specific). The router dynamically selects the top-2 most relevant experts for each input sample based on the learned gating weights, enabling adaptive feature specialization while maintaining overall training stability through load-balancing regularization.

\begin{figure*}[htbp]
    \centering
    \begin{subfigure}[b]{0.33\textwidth}
        \centering
        \includegraphics[width=\textwidth]{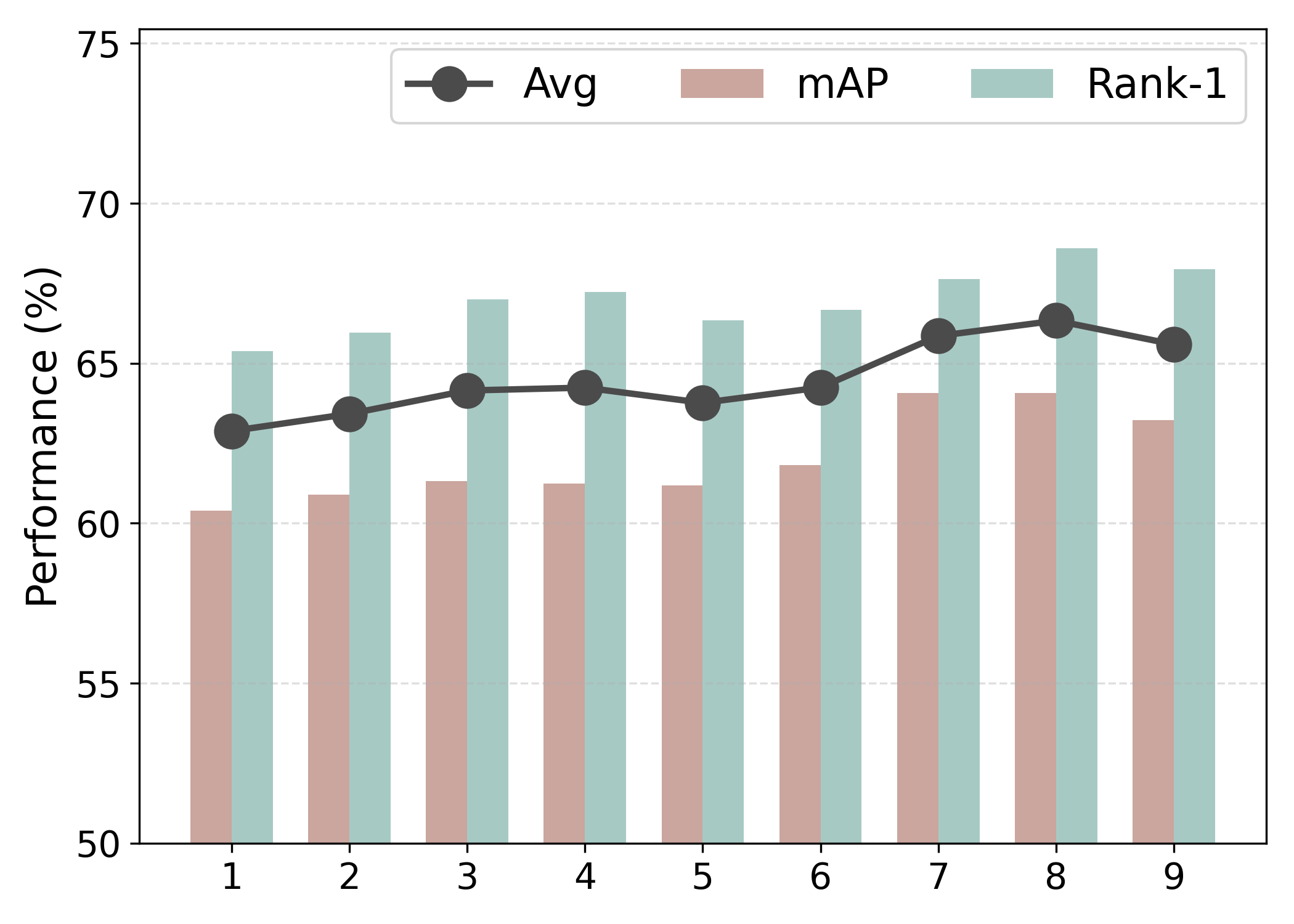}
        \caption{The effect of $E$}
        \label{fig:proto1}
    \end{subfigure}
    \begin{subfigure}[b]{0.33\textwidth}
        \centering
        \includegraphics[width=\textwidth]{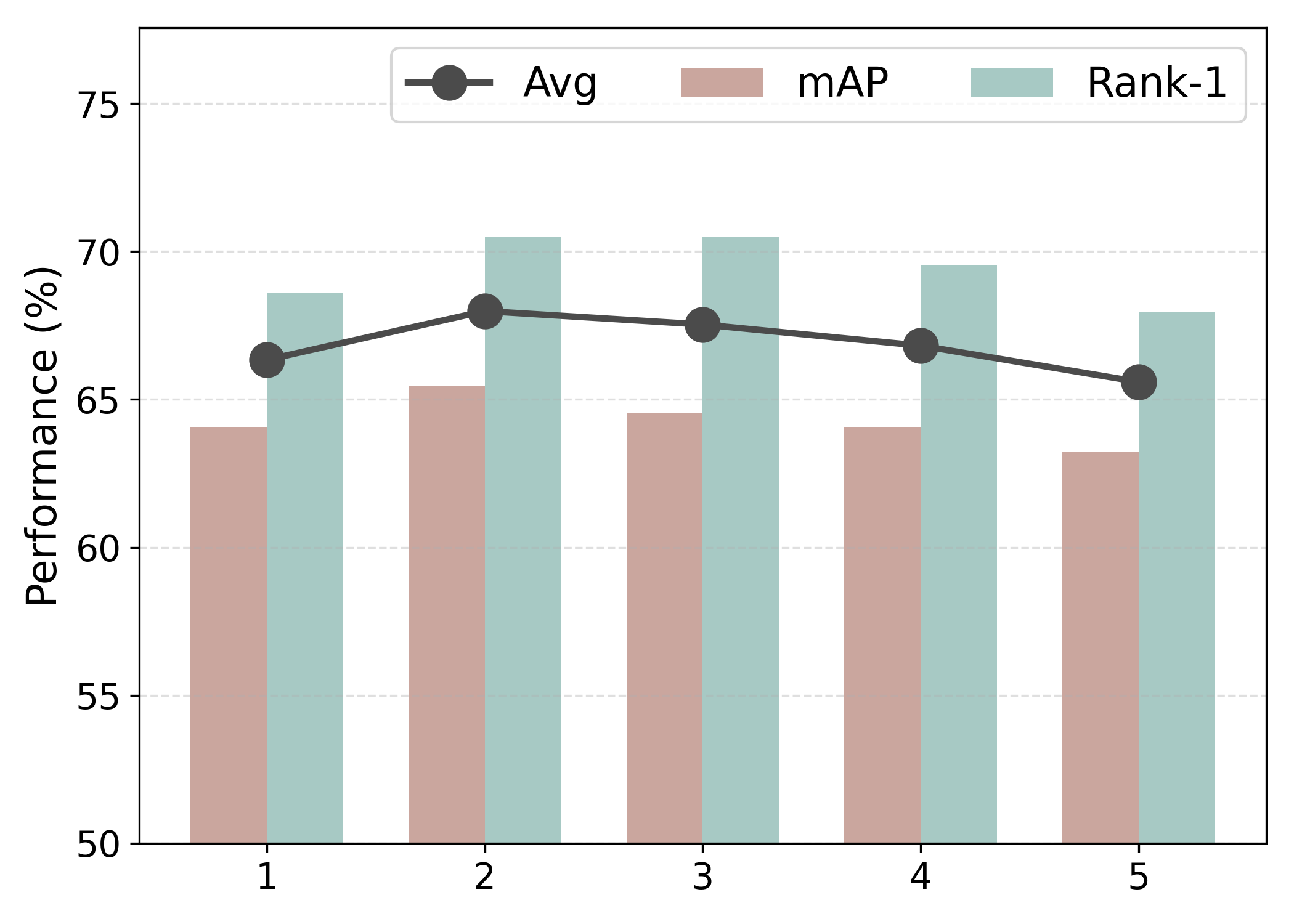}
        \caption{The effect of $k$}
        \label{fig:proto3}
    \end{subfigure}
    \begin{subfigure}[b]{0.33\textwidth}
        \centering
        \includegraphics[width=\textwidth]{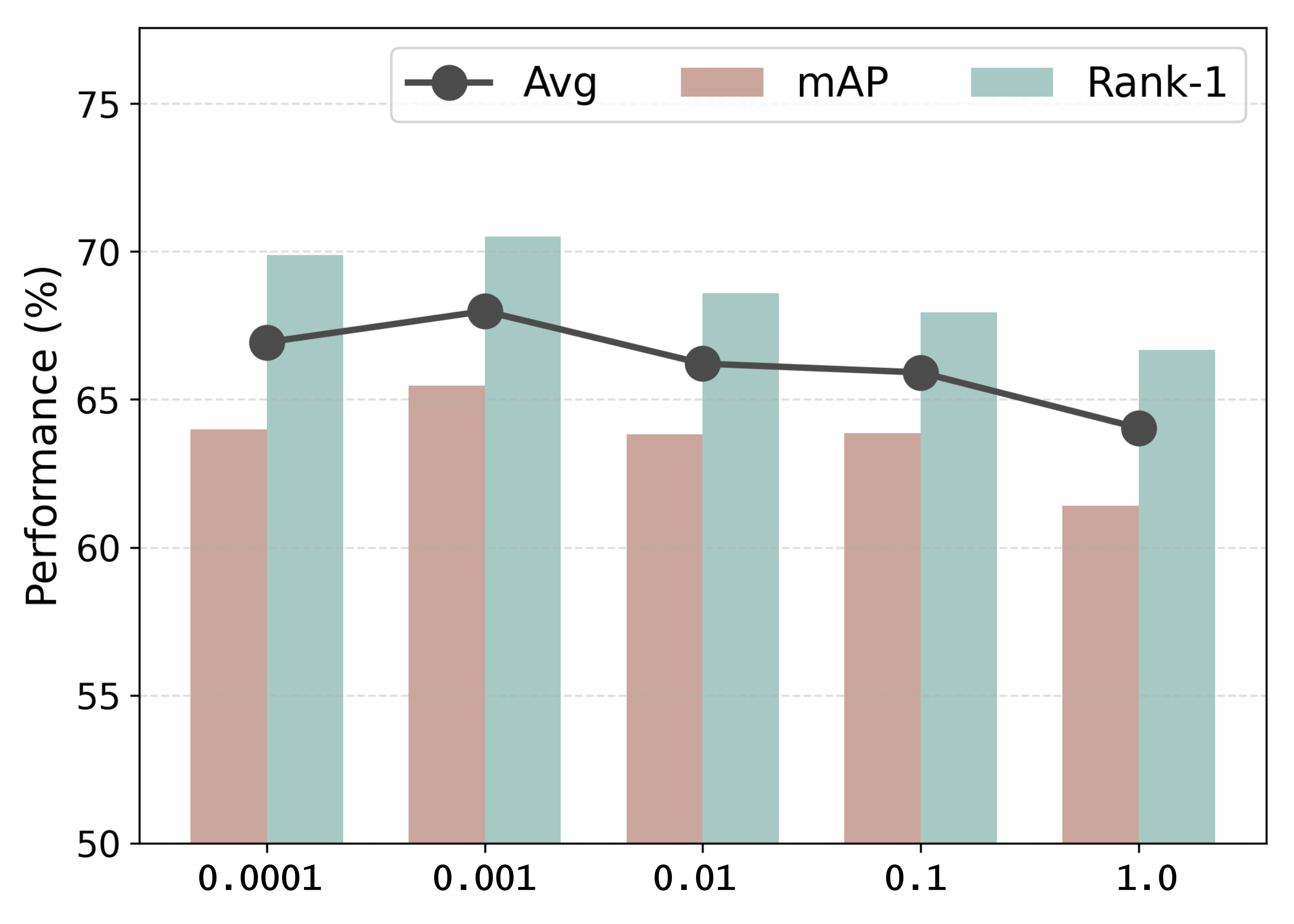}
        \caption{The effect of $\lambda$}
        \label{fig:proto2}
    \end{subfigure}
    \caption{Analysis of hyper-parameters under `ALL' protocol on the CARGO dataset. Rank-1 and mAP and their average are reported (\%). $E$ represents the number of experts. $k$ represents number of selected experts. For further comparison on other protocols, please refer to the supplementary material.}
    \label{fig:parameter}
\end{figure*}

\subsection{Performance}

Tab.~\ref{tab:cargo_full} report ViSA’s performance on CARGO datasets. ViSA achieves 70.51\% Rank-1 and 65.46\% mAP under the ALL protocol, outperforming the previous best method by +1.92\% Rank-1 and +5.27\% mAP. Notably, ViSA also shows substantial improvements in the A$\leftrightarrow$G protocol, achieving 71.28\% Rank-1 and 69.00\% mAP, which demonstrates its effectiveness in bridging the extreme viewpoint gap between aerial and ground cameras. These gains indicate that our expert-driven disentanglement, combined with query-guided local graph fusion, not only enhances global cross-view representations but also effectively exploits fine-grained patch-level cues, enabling more robust and discriminative identity modeling across heterogeneous views.

Performance on other datasets are shown in the supplementary material. On AG-ReID.v2, ViSA achieves the highest mAP across all four evaluation protocols and almost the best Rank-1 accuracy, demonstrating consistent improvements over ViT~\cite{dosovitskiy2020image} and recent AGPReID methods including VDT~\cite{zhang2024view}, V2E~\cite{nguyen2024agreidv2}, and SeCap~\cite{wang2025secap}.

On LAGPeR, ViSA achieves the best mAP under all three protocols and ranks first or second for Rank-1, further confirming the effectiveness of expert-driven disentanglement and local graph-based fusion in capturing complementary identity cues across heterogeneous views. Additional results and comparisons are provided in the supplementary material, demonstrating consistent state-of-the-art performance across diverse aerial-ground settings. 

Overall, ViSA sets a new state-of-the-art on standard AGPReID benchmarks, demonstrating that view-aware semantic alignment provides a powerful and general paradigm for capturing both shared and view-specific identity cues in cross-view person re-identification.

\subsection{Ablation Study}

We conduct extensive ablation studies on the CARGO dataset to evaluate the contribution of each component in ViSA, including the View-Aware Backbone (VAB), Expert-driven Token Generation Module (ETGM), and Dual-branch Local Fusion Module (DLFM). The results are summarized in Tab.~\ref{tab:cargo_ab}. From the results, several observations can be made. (1) Adding DLFM alone to the baseline significantly improves performance (ALL Rank-1: 61.54\% $\to$ 67.31\%, mAP: 53.54\% $\to$ 62.86\%), demonstrating the importance of capturing structural and semantic relations among local patch features. By explicitly modeling local interactions through graph convolution and top-$k$ neighborhood selection, DLFM effectively exploits distributed identity information and enhances discriminability. (2) Removing ETGM results in lower performance (A$\leftrightarrow$G Rank-1: 71.28\% $\to$ 68.09\%) since the model lacks a mechanism to explicitly separate identity cues from viewpoint-dependent variations, which is particularly important in cross-view scenarios. (3) Equipping the model with the VAB further boosts results, especially in challenging cross-view settings. 

Overall, combining all three components achieves the best performance, highlighting the complementary roles of view-aware mechanism, semantic-guided expert queries, and structured local fusion. 

\subsection{Parameter Analysis}

We further conduct a detailed parameter analysis to investigate the impact of key hyperparameters in our ViSA framework, including the number of experts $E$, the number of selected experts $k$, and the balancing coefficient $\lambda$. Results are illustrated in Fig.~\ref{fig:parameter}. We vary the number of experts in the ETGM from 1 to 9. As shown in Fig.~\ref{fig:parameter}(a), the overall performance initially increases with more experts and reaches the best result when $E$ is set to 8. This indicates that introducing multiple experts allows the model to better capture diverse viewpoint-specific patterns. However, an excessive number of experts slightly degrades performance, likely due to redundancy and increased competition among experts. We then analyze the effect of selecting different numbers of active experts during routing. As shown in Fig.~\ref{fig:parameter}(b), the best performance is achieved when two experts are activated per sample. Activating a single expert limits the model’s representational capacity, while selecting too many experts (e.g., more than 3) dilutes expert specialization and introduces noisy feature aggregation. Finally, we analyze the influence of the coefficient $\lambda$, which controls the weight of the load-balancing loss in the MoE routing process. As shown in Fig.~\ref{fig:parameter}(c), setting $\lambda=0.001$ achieves the best performance. A small balancing strength encourages sufficient diversity among experts while maintaining focus on discriminative learning. Conversely, a large $\lambda$ enforces overly uniform expert utilization, suppressing specialization and leading to degraded retrieval accuracy. 

\subsection{Visualization Analysis}

\begin{figure}[t]
    \centering
    \includegraphics[width=\linewidth]{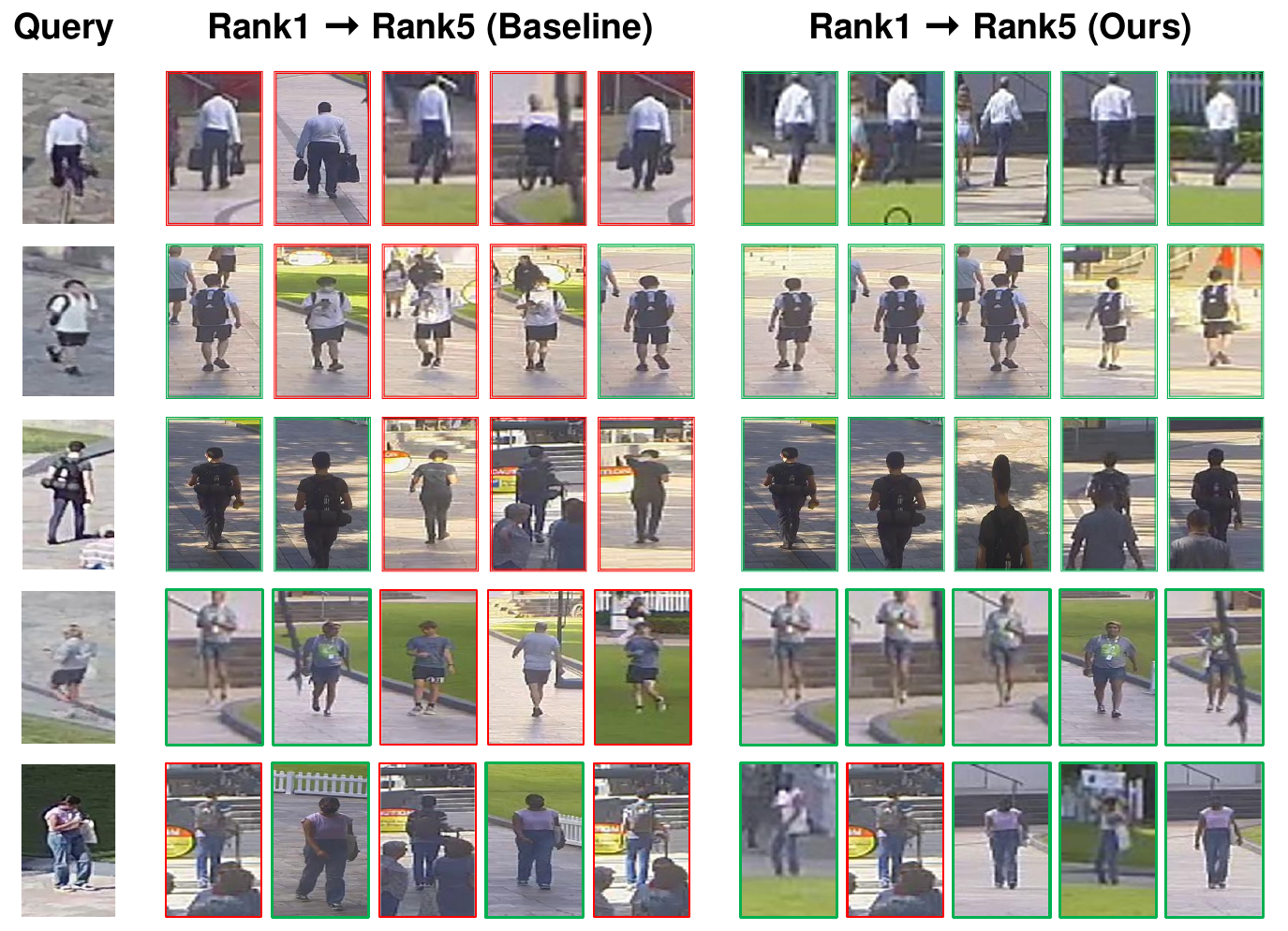}
    \caption{
        Comparison of several retrieval visualizations on AG-ReID.v2 dataset under A$\to$G protocol. Red and green boxes represent wrong and correct matchings, respectively.
    }
    \vspace{-3mm}
    \label{fig:supp_rank5_vis}
\end{figure}

To further validate the effectiveness of the view-aware semantic alignment strategy, we conducted a retrieval visualization analysis on the AG-ReID.v2 dataset under A$\to$G protocol (Fig.~\ref{fig:supp_rank5_vis}). Unlike the baseline, ViSA consistently matches identities across viewpoints, suggesting that the proposed view-aware alignment mitigates local alignment degradation and achieves semantic-level consistency.

\section{Conclusion}

In this work, we focus on the challenging problem of AGPReID and highlight that existing view-invariant paradigms may collapse into part-level alignment, thereby failing to fully exploit discriminative identity cues tied to viewpoint variations. To tackle this issue, we propose a view-aware semantic alignment framework named ViSA that achieves consistent cross-view perception by constructing adaptive viewpoint-specific experts and aligning their responsive local regions through graph reasoning. Extensive experiments on AG-ReID.v2, CARGO, and LAGPeR demonstrate that ViSA consistently surpasses state-of-the-art methods, achieving up to a 10.06\% mAP improvement under the challenging cross-view protocol. These results verify the effectiveness of jointly modeling view-invariant and view-specific cues for robust aerial-ground person re-identification.

\section*{Acknowledgments}
{This work was supported by the National Natural Science Foundation of China (U22A2095, 62506393), the Project of Guangdong Provincial Key Laboratory of Information Security Technology (2023B1212060026), the Guangdong Basic and Applied Basic Research Foundation (2026A1515011438), and the Postdoctoral Fellowship Program and the China Postdoctoral Science Foundation (GZC20252314).}

{
    \small
    \bibliographystyle{ieeenat_fullname}
    \bibliography{main}
}

\clearpage
\setcounter{page}{1}
\maketitlesupplementary

In this supplementary material, we provide additional analyses and results to further evaluate our proposed ViSA. Specifically, we first present extended qualitative visualizations, including additional retrieval examples on CARGO and t-SNE comparisons. We then report comprehensive quantitative results by providing full performance tables on AG-ReID.v2 and LAGPeR under all evaluation settings. Finally, we conduct detailed hyperparameter analyses, examining the effects of the number of experts, the number of selected experts, and the balancing coefficient across different cross-view protocols.

\section{Visual Analysis}

\subsection{Retrieval Visualization}

\begin{figure}[t]
    \centering
    \includegraphics[width=\linewidth]{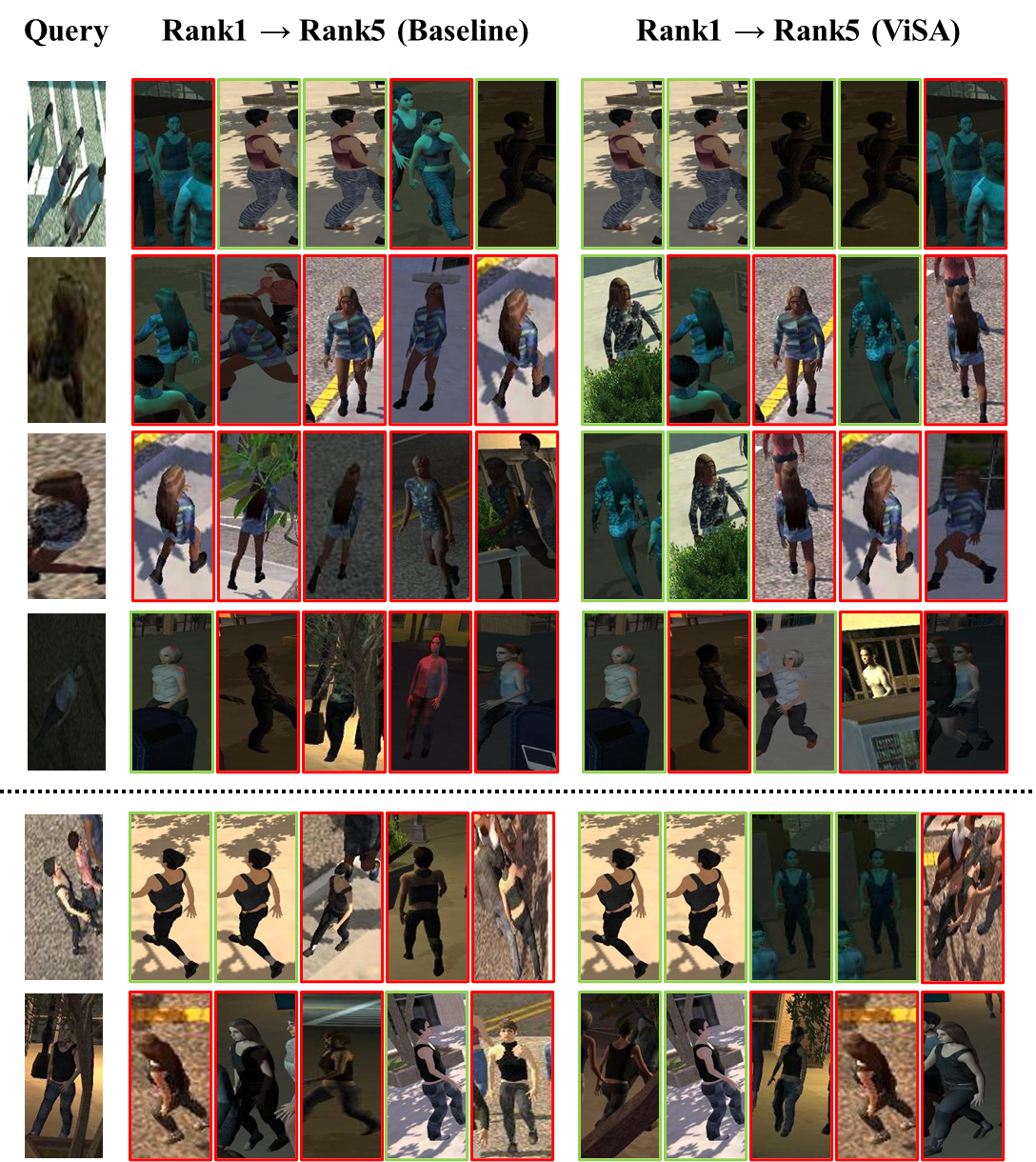}
    \caption{
        {Comparison of several retrieval visualizations on CARGO dataset under A$\to$G protocol and ALL protocol. Red and green boxes represent wrong and correct matchings, respectively. The top five retrieved results are shown.}
    }
    \label{fig:supp_rank5_vis_cargo}
\end{figure}

In addition to the retrieval visualization on the AG-ReID.v2 dataset shown in Fig.~\ref{fig:supp_rank5_vis}, we further provide retrieval examples on the CARGO dataset under both the A$\leftrightarrow$G protocol and the ALL protocol. As illustrated in Fig.~\ref{fig:supp_rank5_vis_cargo}, our method consistently retrieves correct cross-view matches across different evaluation settings, demonstrating strong robustness and generalization. These results further verify the effectiveness of our approach in handling diverse viewpoint transitions across datasets and protocols.

\subsection{Feature Visualization}

\begin{figure}[t]
    \centering
    {\includegraphics[width=0.4\linewidth]{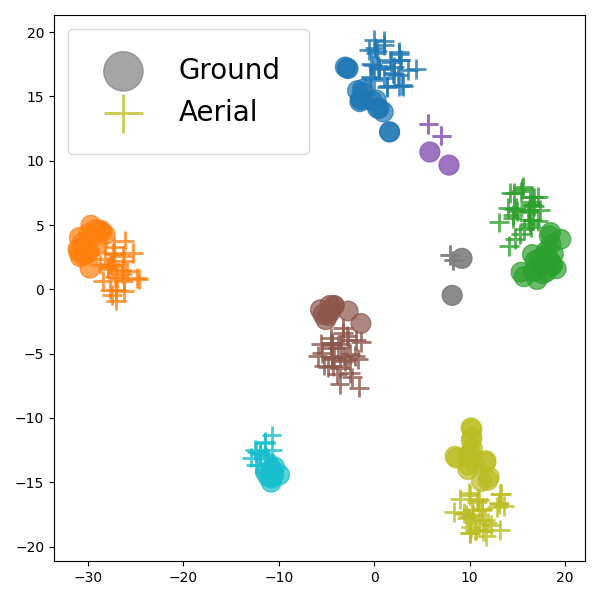}}
    {\includegraphics[width=0.4\linewidth]{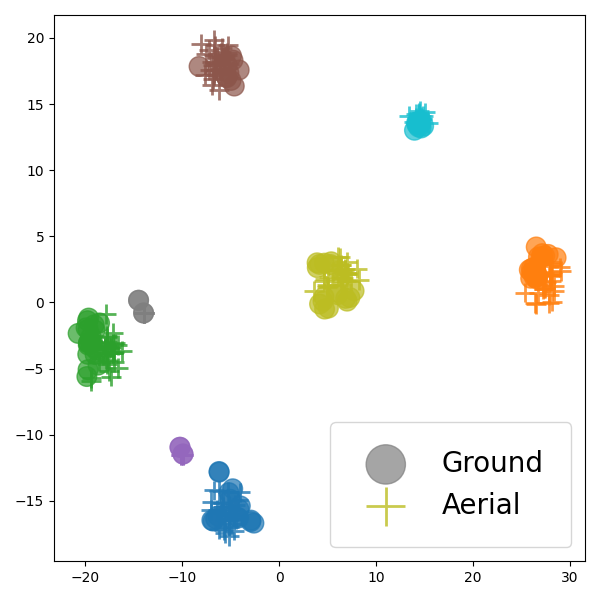}}
    \caption{{t-SNE visualization. \textbf{Left:} Baseline. \textbf{Right:} ViSA. Colors indicate different classes and shapes denote different views.}}
    \label{fig:t-SNE}
\end{figure}

To further investigate the effectiveness of ViSA in cross-view representation learning, we visualize the learned embeddings using \textit{t}-SNE. As shown in Fig.~\ref{fig:t-SNE}, we compare the feature distributions of the baseline model and ViSA. The reduced discrepancy between ground and aerial features indicates that ViSA effectively mitigates cross-view domain gaps and learns more discriminative, view-invariant representations.

\section{Performance}





\begin{table*}[t]
    \centering
    \caption{Performance comparison on AG-ReID.v2 dataset. C represents CCTV, W represents wearable devices and A represents aerial views. The best and second-best results are highlighted in \textbf{bold} and \underline{underline}, respectively.}
        \begin{tabular}{lccccccccc}
        \toprule
        \multirow{2}{*}{Method} & \multirow{2}{*}{Venue} & \multicolumn{2}{c}{A$\to$C} & \multicolumn{2}{c}{A$\to$W} & \multicolumn{2}{c}{C$\to$A} & \multicolumn{2}{c}{W$\to$A}  \\
         \cmidrule(lr){3-4}  \cmidrule(lr){5-6} \cmidrule(lr){7-8} \cmidrule(lr){9-10}
        & & Rank-1 & mAP & Rank-1 & mAP & Rank-1 & mAP & Rank-1 & mAP \\
        \midrule
        Swin \cite{liu2021swin} & ICCV'21 & 68.76 & 57.66 & 68.49 & 56.15 & 68.80 & 57.70 & 64.40 & 53.90 \\
        HRNet-18 \cite{wang2020deep} & PAMI'20 & 75.21 & 65.07 & 76.26 & 66.17 & 76.25 & 66.16 & 76.25 & 66.17 \\
        SwinV2 \cite{liu2022swin} & CVPR'22 & 76.44 & 66.09 & 80.08 & 69.09 & 77.11 & 62.14 & 74.53 & 65.61 \\
        MGN(R50) \cite{2018Learning} & ACM MM'18 & 82.09 & 70.17 & 88.14 &  78.66 &  84.21 & 72.41 & 84.06 & 73.73 \\
        BoT(R50) \cite{luo2019bag} & CVPR'19 & 80.73 & 71.49 & 86.06 & 75.98 & 79.46 & 69.67 & 82.69 & 72.41 \\
        SBS(R50) \cite{he2023fastreid} & ACM MM'23 & 81.96 & 72.04 & 88.14 & 78.94 & 84.10 & 73.89 & 84.66 & 75.01 \\
        \midrule
        BoT(ViT) \cite{luo2019bag} & CVPR'19 & 85.40 & 77.03 & 89.77 & 80.48 & 84.65 & 75.90 & 84.65 & 75.90 \\
        ViT \cite{dosovitskiy2020image} & ICLR'21 & 85.40 & 77.03 & 89.77 & 80.48 & 84.65 & 75.90 & 84.27 & 76.59 \\
        TransReID \cite{he2021transreid} & ICCV'21 & 88.00 & \underline{81.40} & 90.40 & 84.50 & 87.60 & \underline{80.10} & 87.70 & \underline{81.10} \\
        FusionReID \cite{wang2025unity} & T-ITS'25 & 86.70 & 80.70 & 89.70 & 84.20 & 87.90 & 80.00 & 86.50 & 80.90 \\
        CLIP-ReID \cite{li2023clip} & AAAI'23 & 85.36 & 79.79 & 89.14 & 84.23 & 85.64 & 79.08 & 86.50 & 79.55 \\
        PCL-CLIP \cite{li2023prototypical} & CoRR'23 & 79.80 & 72.20 & 87.14 & 77.70 & 81.12 & 72.40 & 84.19 & 73.89 \\
        \midrule
        Explain \cite{nguyen2023agreid} & ICME'23 & 87.70 & 79.00 & \textbf{93.67} & 83.14 & 87.35 & 78.24 & 87.73 & 79.08 \\
        VDT \cite{zhang2024view} & CVPR'24 & 86.46 & 79.13 & 90.00 & 82.21 & 86.14 & 78.12 & 85.26 & 78.52 \\
        V2E(ViT) \cite{nguyen2024agreidv2} & TIFS'24 & \underline{88.77} & 80.72 & \underline{93.62} & \underline{84.85} & 87.86 & 78.51 & \underline{88.61} & 80.11 \\
        SeCap \cite{wang2025secap} & CVPR'25 & 88.12 & 80.84 & 91.44 & 84.01 & \underline{88.24} & 79.99 & 87.56 & 80.15 \\
        \midrule
        \rowcolor{gray!15}
        ViSA & - & \textbf{89.43} & \textbf{83.61} & {91.63} & \textbf{85.99} & \textbf{88.57} & \textbf{82.32} & \textbf{89.23} & \textbf{83.10} \\
        \bottomrule    
        \end{tabular}
    \label{tab:agreidv2_full}
\end{table*}

\begin{table*}[t]
    \centering
    \caption{Performance comparison on LAGPeR. The best and second-best results are highlighted in \textbf{bold} and \underline{underline}, respectively. 'A$\leftrightarrow$G' denotes that the Aerial view is the query, 'G$\leftrightarrow$A' denotes that the Ground view is the query, and 'G$\leftrightarrow$A+G' indicates that the gallery contains images from both the Aerial and Ground view. CLIP-ReID$^*$ indicates using OLP and SIE in Clip-ReID. MIP$\dagger$ represents the re-implementation for the AGPReID. Explain$\ddagger$ indicates removing the attributes branch of the Explain method.}
        \begin{tabular}{lccccccccccc}
        \toprule
        \multirow{2}{*}{Method} & \multirow{2}{*}{Venue} & \multicolumn{2}{c}{A$\leftrightarrow$G} & \multicolumn{2}{c}{G$\leftrightarrow$A} & \multicolumn{2}{c}{G$\leftrightarrow$A+G} & \multicolumn{2}{c}{Average} \\
         \cmidrule(lr){3-4}  \cmidrule(lr){5-6} \cmidrule(lr){7-8} \cmidrule(lr){9-10}
        & & Rank-1 & mAP & Rank-1 & mAP & Rank-1 & mAP & Rank-1 & mAP \\
        \midrule
        \midrule
        ViT \cite{dosovitskiy2020image} & ICLR'21  & 38.67 & 27.25 & 32.04 & 30.69 & 18.88 & 15.31 & 29.86 & 24.42 \\
        TransReID \cite{he2021transreid} & ICCV'21 & 38.80 & 28.80 & 33.00 & 32.10 & 22.90 & 18.80 & 31.57 & 26.57 \\
        CLIP-ReID \cite{li2023clip} & AAAI'23 & 24.40 & 17.60 & 21.30 & 20.80 & 12.30 & 10.20 & 19.33 & 16.20 \\
       CLIP-ReID$^*$ & AAAI'23 & 23.10 & 17.50 & 20.00 & 20.30 & 9.00 & 8.40 & 17.37 & 15.40 \\
        MIP$\dagger$ \cite{Wu2024Enhancing} & ICMR'24 & 39.30 & 29.30 & 33.90 & 32.60 & 21.00 & 17.30 & 31.40 & 26.40 \\
        \midrule
        Explain$\ddagger$ \cite{nguyen2023agreid} & ICME'23 & 40.48 & 28.89 & 32.96 & 31.91 & 22.03 & 17.89 & 31.82 & 26.23 \\
        VDT \cite{zhang2024view} & CVPR'24 & 40.15 & 28.97 & 33.55 & 31.98 & 19.50 & 16.45 & 31.07 & 25.80 \\
        SeCap \cite{wang2025secap} & CVPR'25 & \textbf{41.79} & \underline{30.37} & \textbf{35.26} & \underline{33.42} & \underline{24.39} & \underline{19.24} & \underline{33.81} & \underline{27.68} \\
        \midrule
        \rowcolor{gray!15}
        ViSA & - & \underline{41.37} & \textbf{30.42} & \underline{35.06} & \textbf{33.59} & \textbf{25.61} & \textbf{20.31} & \textbf{34.01} & \textbf{28.11} \\
        \bottomrule
        \end{tabular}   
    \label{tab:lagper}
\end{table*}

Tab.~\ref{tab:agreidv2_full} and Tab.~\ref{tab:lagper} report the performance of ViSA on AG-ReID.v2 and LAGPeR, respectively. On AG-ReID.v2, ViSA achieves the highest mAP across all methods and sets a new state of the art in every evaluation setting except the A$\rightarrow$W protocol, where the viewpoint variation is relatively minor. We further observe that in this less challenging A$\rightarrow$W scenario, methods such as Explain and V2E obtain slightly better results, largely benefiting from the use of attribute annotations. When compared only with approaches that do not rely on attribute labels, ViSA still delivers the best performance. On the LAGPeR dataset, ViSA achieves improvements of up to 1.0\% in mAP and 1.22\% in Rank-1 accuracy, demonstrating its strong generalization ability under larger cross-view variations.

\section{Parameter Analysis}

\begin{figure*}[t]
    \centering
    \begin{subfigure}[b]{0.33\textwidth}
        \includegraphics[width=\textwidth]{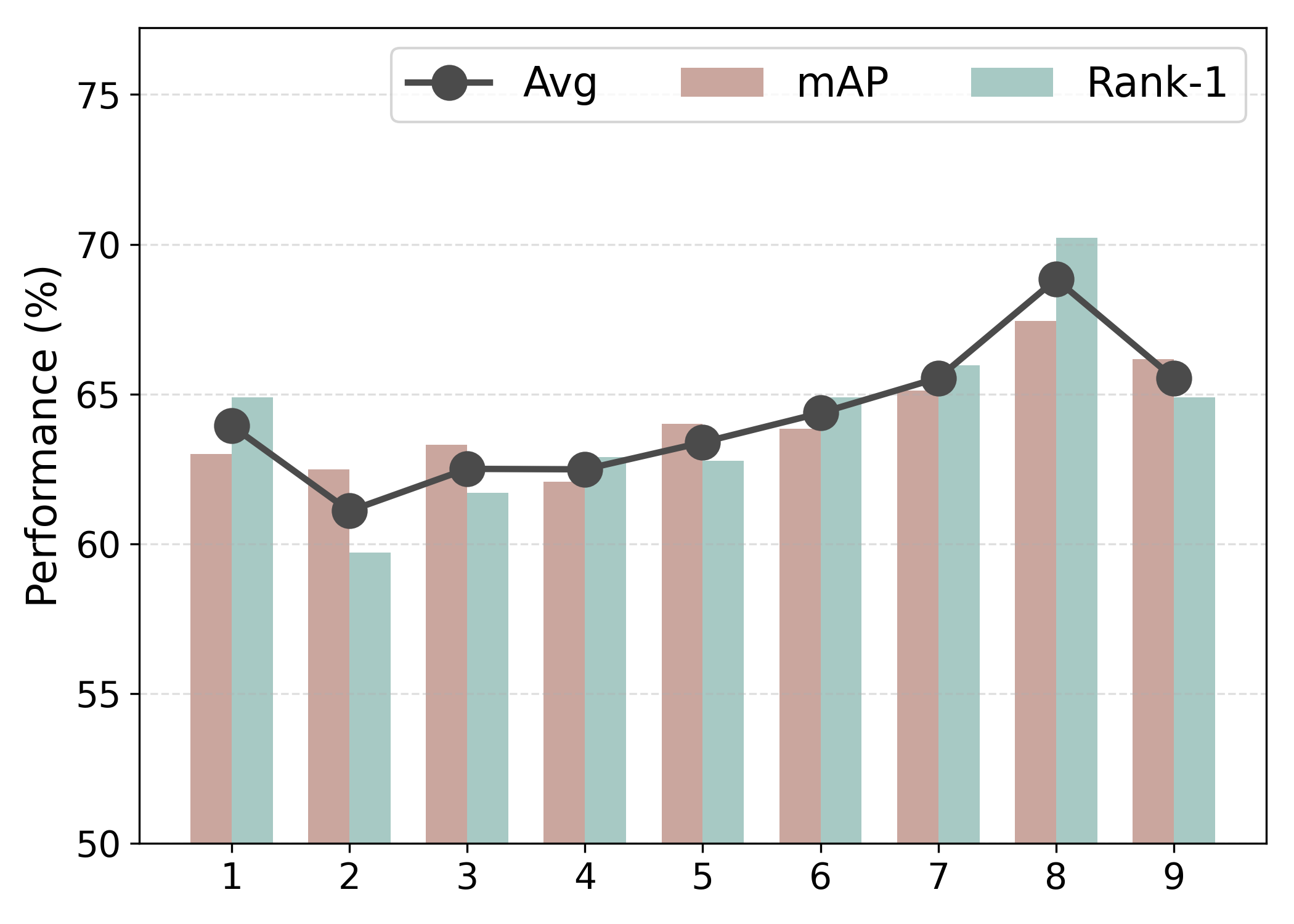}
        \caption{E: A$\leftrightarrow$G}
        \label{fig:a}
    \end{subfigure}
    \begin{subfigure}[b]{0.33\textwidth}
        \includegraphics[width=\textwidth]{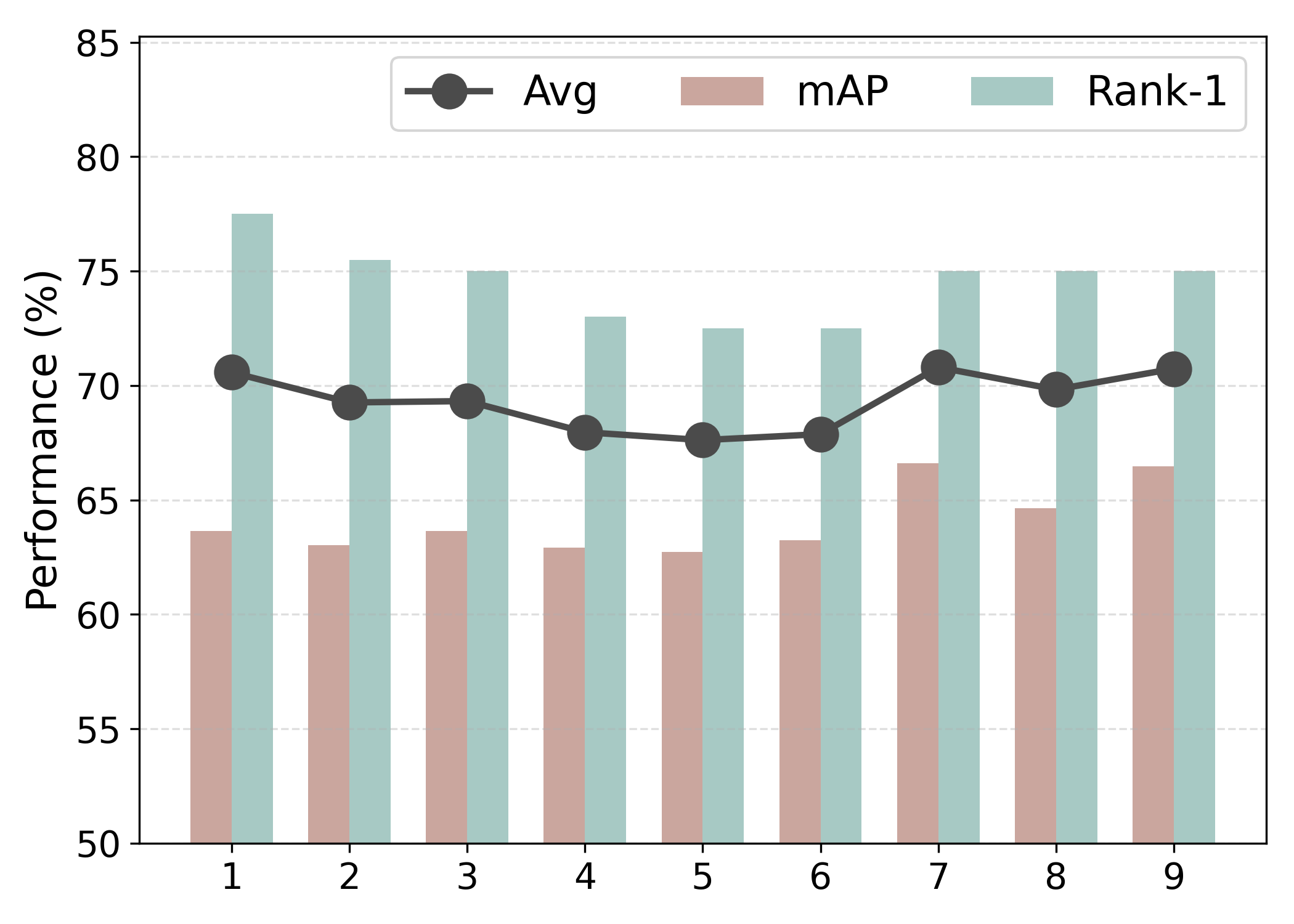}
        \caption{E: A$\leftrightarrow$A}
    \end{subfigure}
    \begin{subfigure}[b]{0.33\textwidth}
        \includegraphics[width=\textwidth]{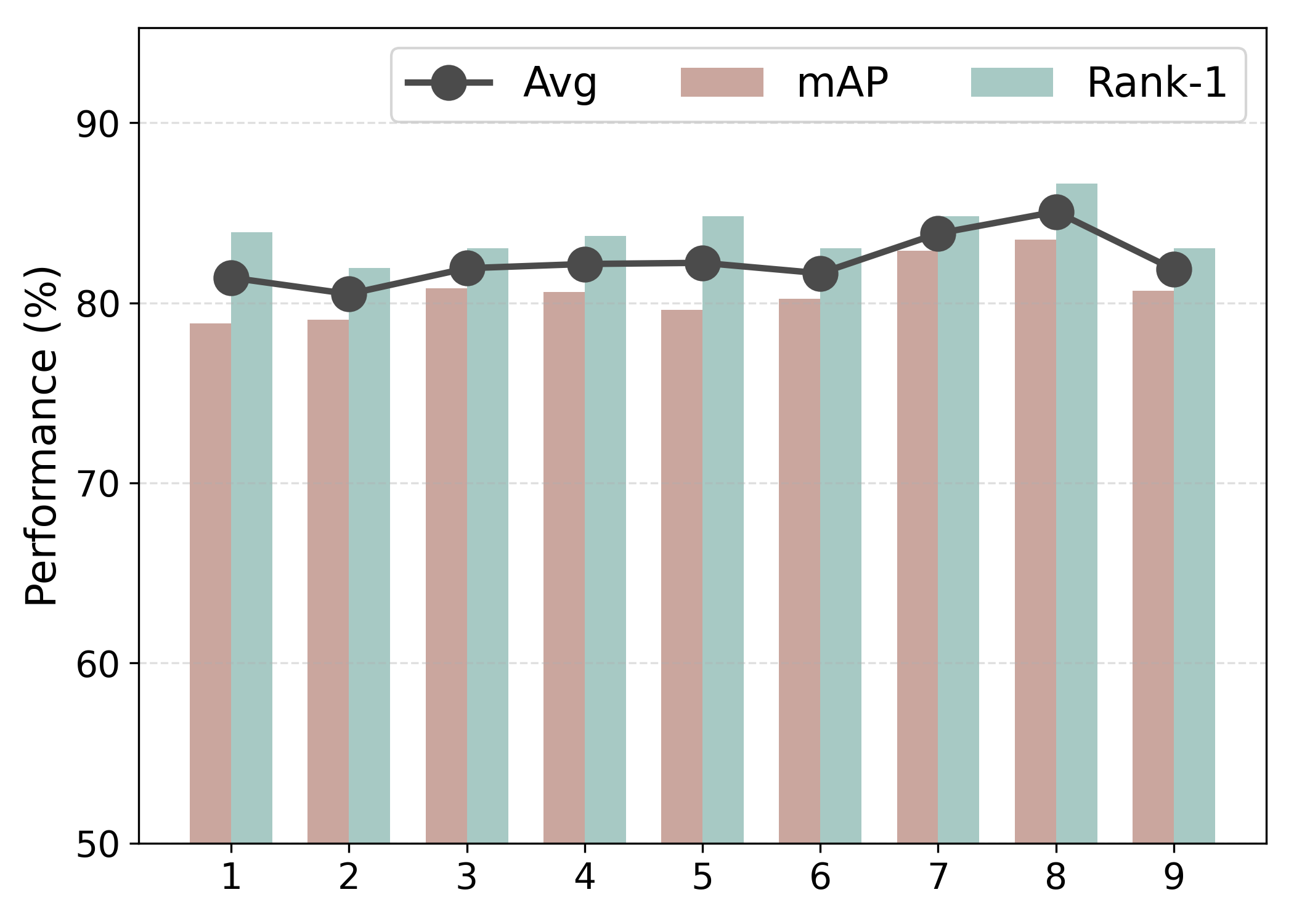}
        \caption{E: G$\leftrightarrow$G}
        \label{fig:c}
    \end{subfigure}

    \begin{subfigure}[b]{0.33\textwidth}
        \includegraphics[width=\textwidth]{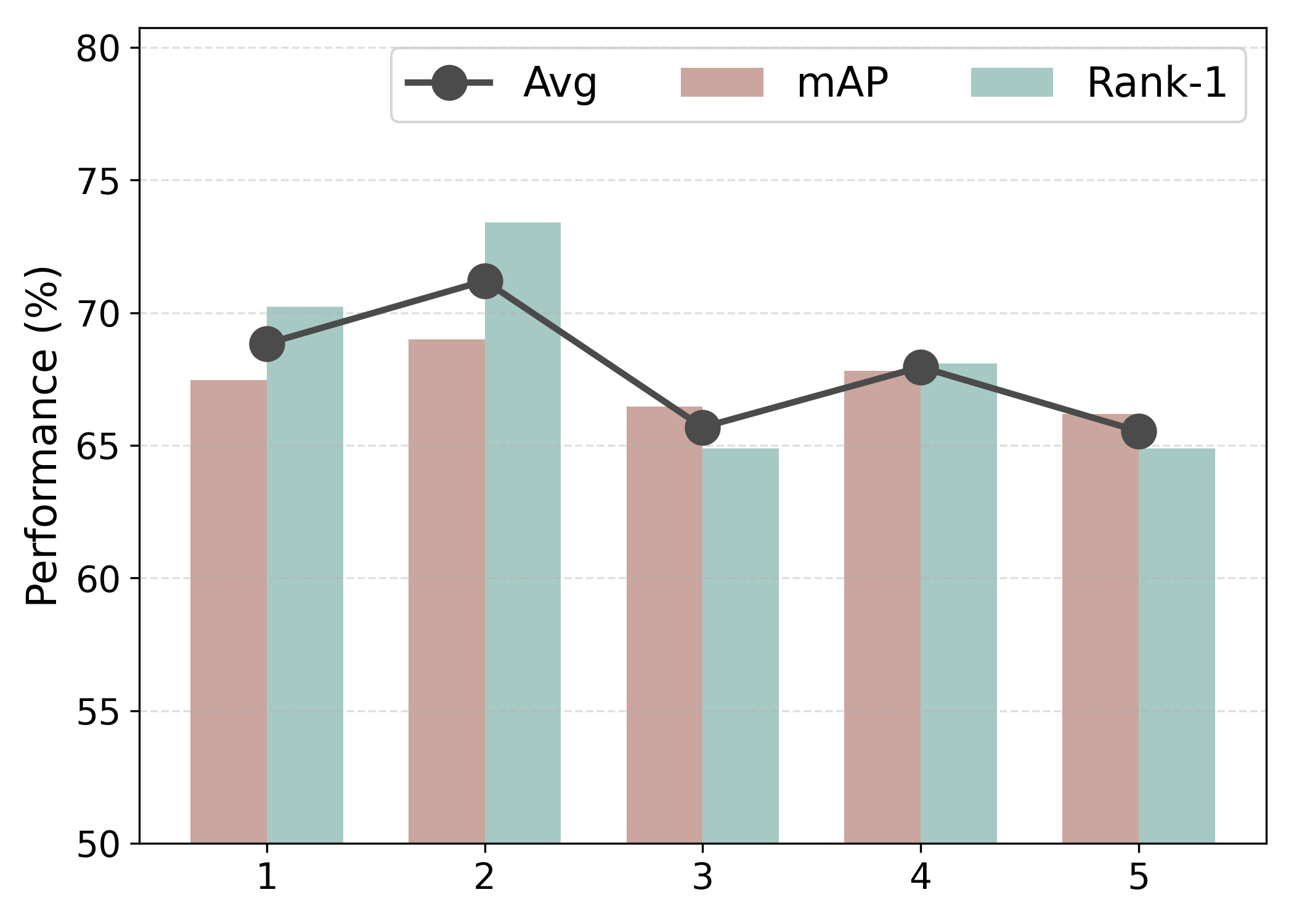}
        \caption{k: A$\leftrightarrow$G}
        \label{fig:d}
    \end{subfigure}
    \begin{subfigure}[b]{0.33\textwidth}
        \includegraphics[width=\textwidth]{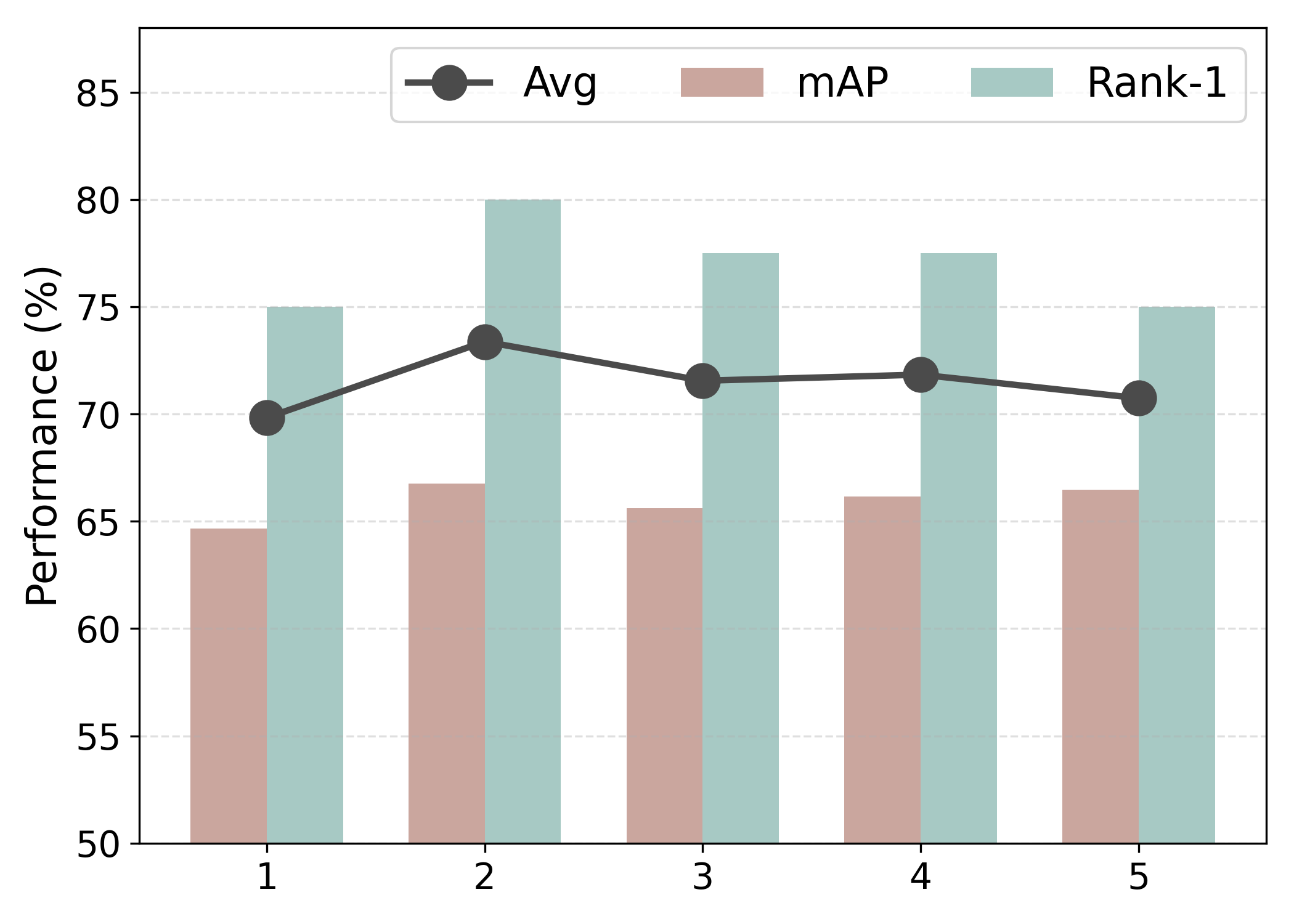}
        \caption{k: A$\leftrightarrow$A}
    \end{subfigure}
    \begin{subfigure}[b]{0.33\textwidth}
        \includegraphics[width=\textwidth]{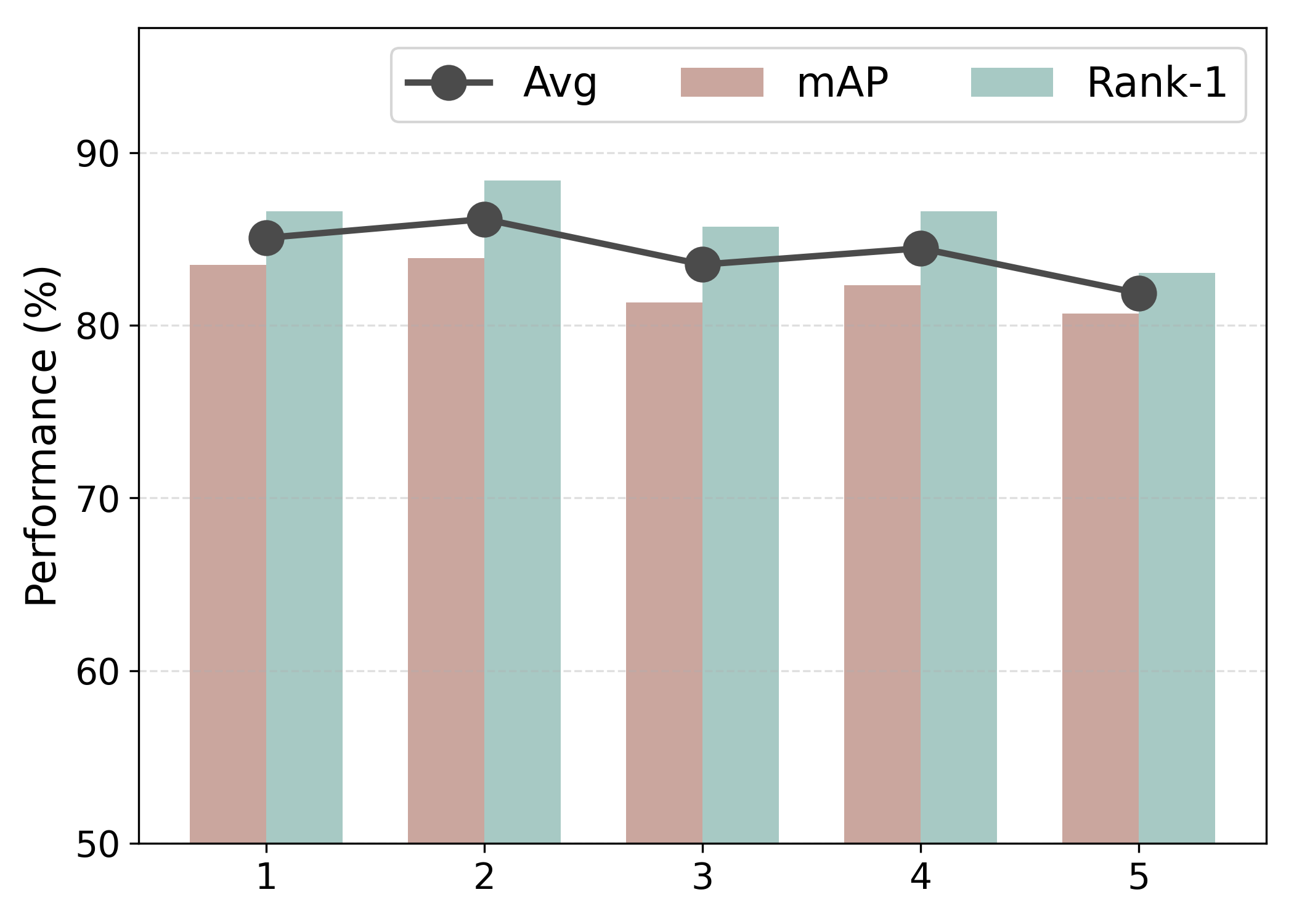}
        \caption{k: G$\leftrightarrow$G}
        \label{fig:f}
    \end{subfigure}

    \begin{subfigure}[b]{0.33\textwidth}
        \includegraphics[width=\textwidth]{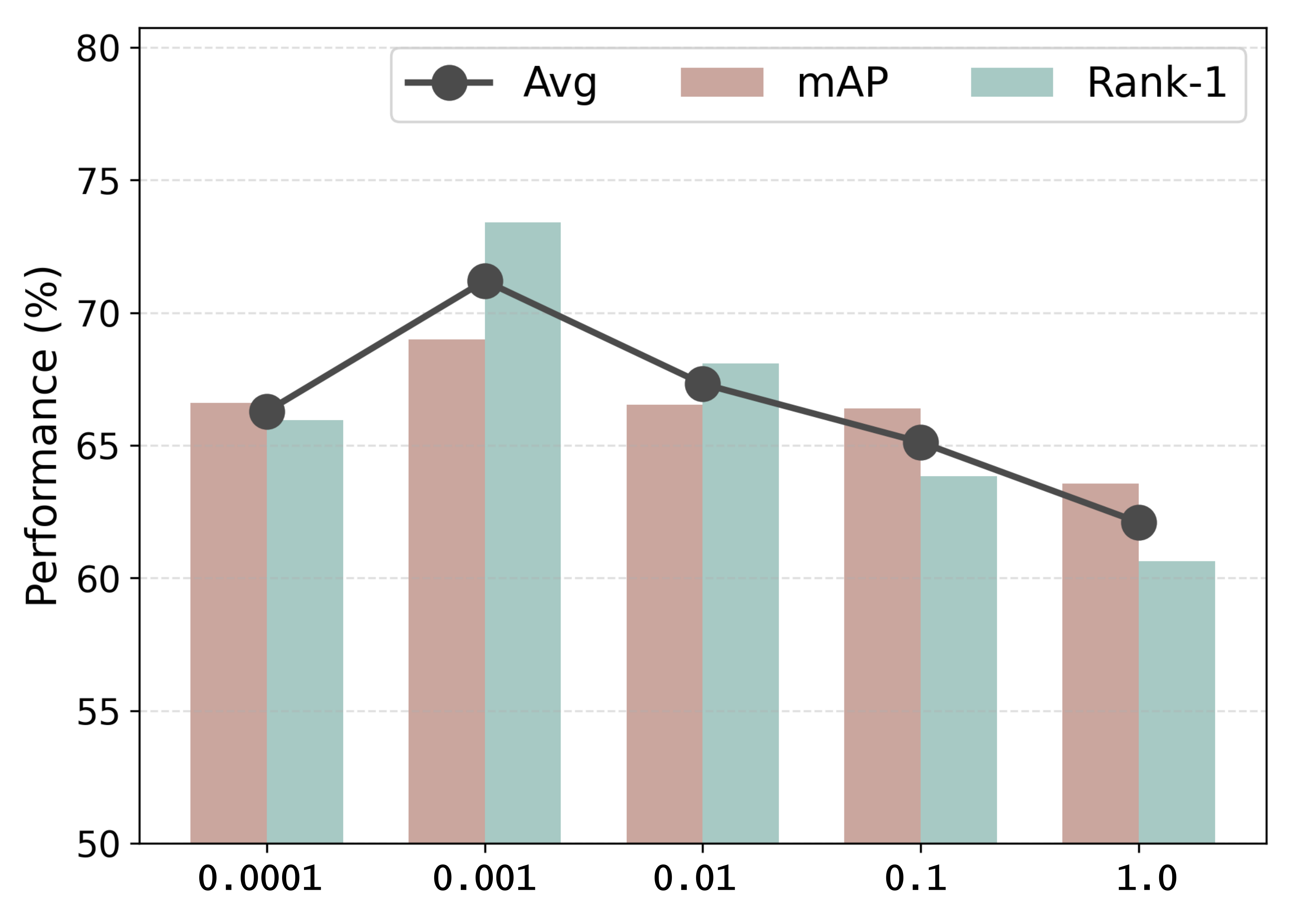}
        \caption{$\lambda$: A$\leftrightarrow$G}
        \label{fig:g}
    \end{subfigure}
    \begin{subfigure}[b]{0.33\textwidth}
        \includegraphics[width=\textwidth]{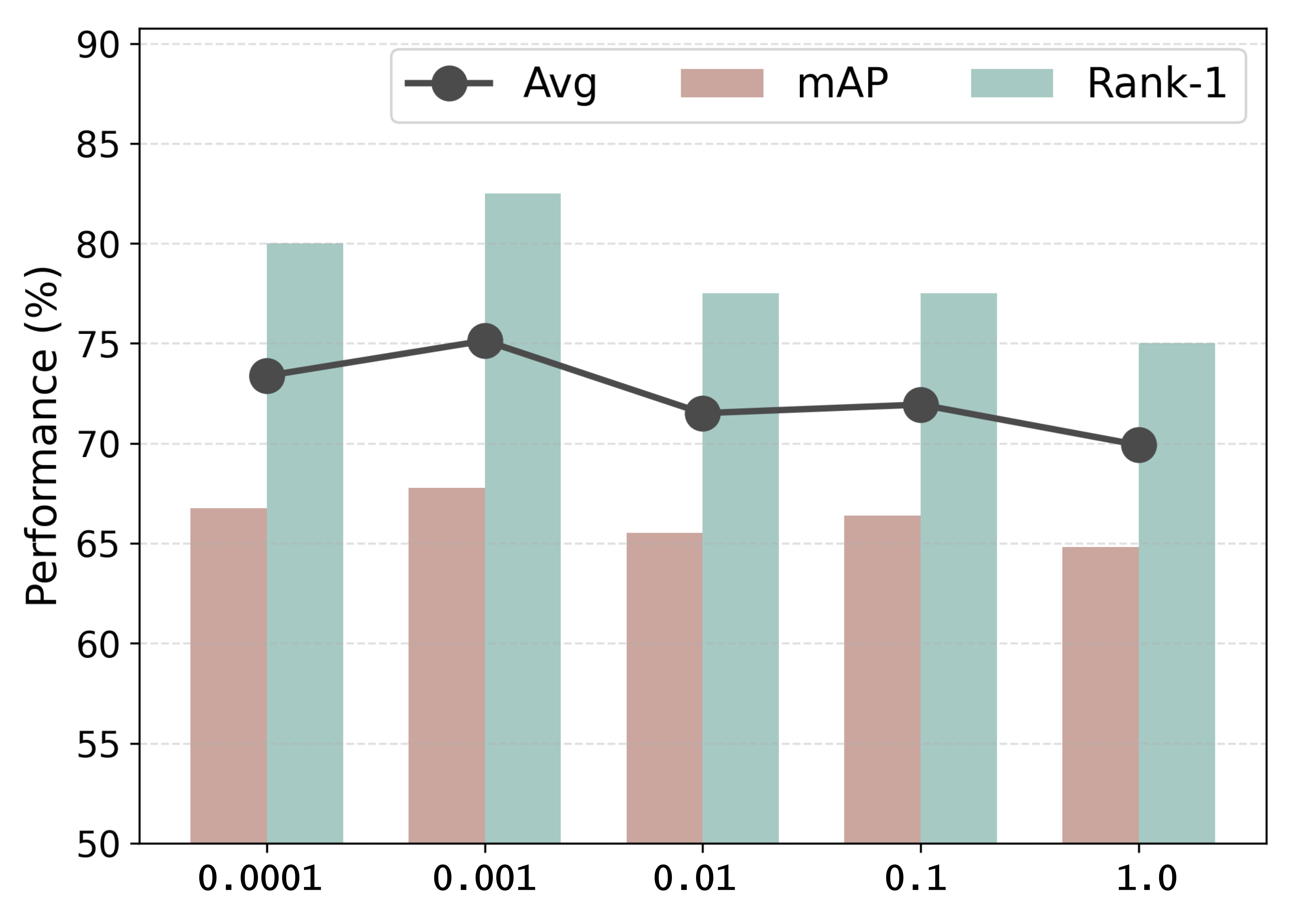}
        \caption{$\lambda$: A$\leftrightarrow$A}
    \end{subfigure}
    \begin{subfigure}[b]{0.33\textwidth}
        \includegraphics[width=\textwidth]{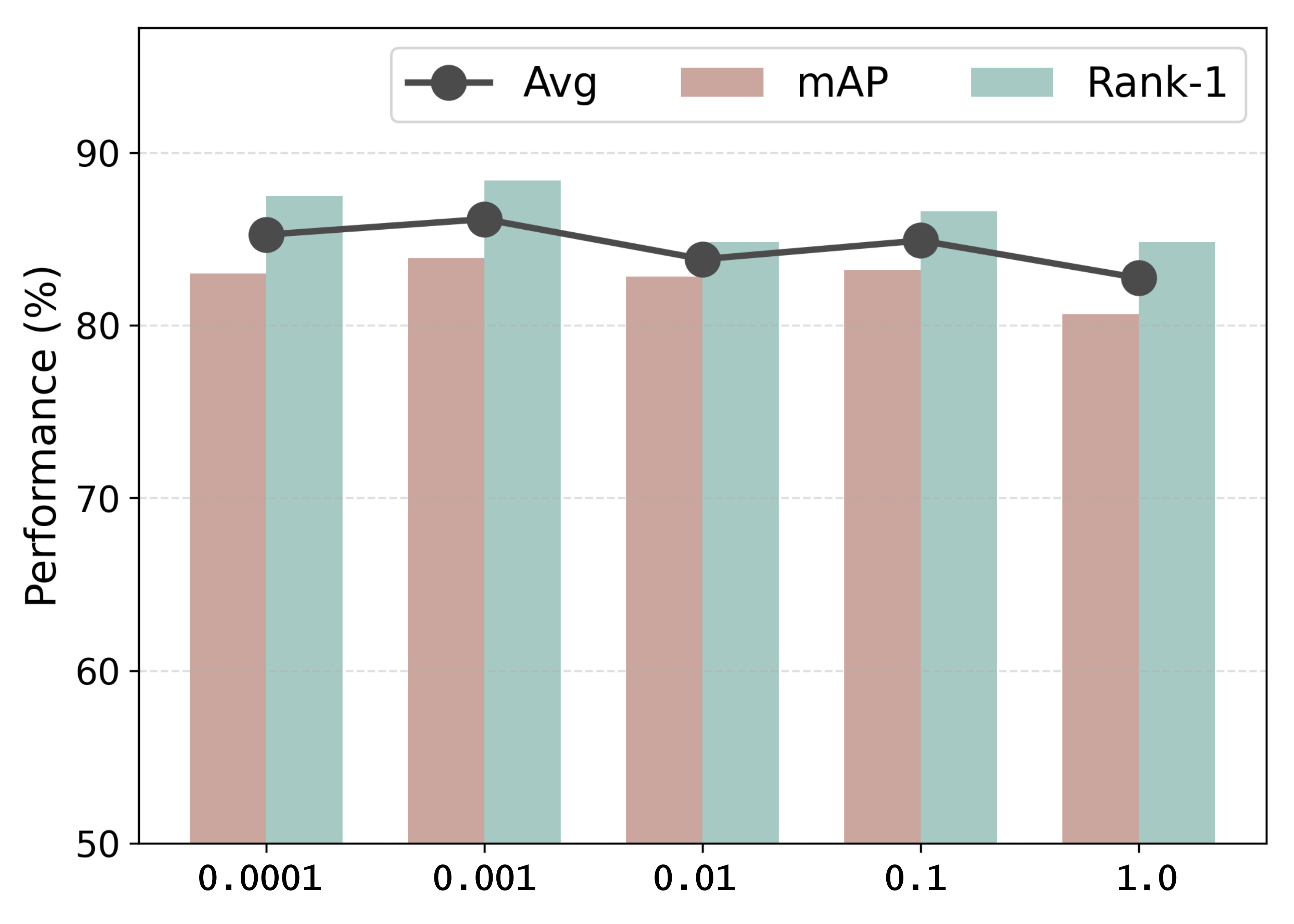}
        \caption{$\lambda$: G$\leftrightarrow$G}
        \label{fig:i}
    \end{subfigure}

    \caption{{Analysis of $E$, $k$, and $\lambda$ across different protocols on the CARGO dataset. Rank-1 and mAP and their average are reported (\%). $E$ represents the number of experts. $k$ represents number of selected experts. $\lambda$ represents the balancing coefficient.}}
    \label{fig:hyper_supp}
\end{figure*}

We further analyze the impact of hyperparameters under different evaluation protocols. In the study of the number of experts $E$, we fix the number of selected expert to 1 (setting $k=1$) to ensure controlled comparisons. As shown in Fig.~\ref{fig:a} $\sim$ Fig.~\ref{fig:c}, the performance first increases and then decreases as $E$ grows. Under the A$\leftrightarrow$G and G$\leftrightarrow$G protocols, the best results are achieved when $E=8$, while under the A$\leftrightarrow$A protocol, the optimal performance is obtained with $E=7$. 

Similarly, as shown in Fig.~\ref{fig:d} $\sim$ Fig.~\ref{fig:f} when analyzing the number of selected experts $k$, we observe a clear rise-then-drop trend. The performance peaks at $k=2$, indicating that activating a small but diverse subset of experts provides the best balance between specialization and generalization.

For the balancing coefficient $\lambda$, we compare values $\{0.0001,0.001,0.01,0.1,1.0\}$ across all protocols. As shown in Fig.~\ref{fig:g} $\sim$ Fig.~\ref{fig:i}, setting $\lambda=0.001$ consistently achieves the best performance on all three protocols, indicating that a moderate regularization strength is essential for maintaining stable expert utilization without suppressing discriminative specialization.

\begin{table}[t]
    \centering
    \caption{Efficiency comparison under ALL protocol.}
\small
        \begin{tabular}{lcccccccc}
        \toprule
        Method & mAP($\uparrow$) & Params($\downarrow$) & GFLOPs($\downarrow$) & FPS($\uparrow$) \\
        \midrule
        SeCap & 58.94\% & 130.88M & 38.81 & 276.66  \\
        ViSA & 69.00\% & 271.77M & 37.53 & 429.05 \\
        \midrule
        Rel. & \textbf{{+10.06\%}} & {+107.6\%} & \textbf{{-3.3\%}} & \textbf{{+55.1\%}}  \\
        \bottomrule    
        \end{tabular}
    \label{tab:cargo_eff}
\end{table}

\section{Efficiency}
ViSA achieves strong accuracy and efficiency despite introducing additional parameters (+107.6\%).
As shown in Tab.~\ref{tab:cargo_eff}, ViSA enhances mAP by 10.06\% while reducing GFLOPs by 3.3\% and increasing inference speed by 55.1\% FPS.

\end{document}